\documentclass[sigconf,nonacm]{acmart}

\usepackage{enumitem}
\AtBeginDocument{%
  \providecommand\BibTeX{{%
    \normalfont B\kern-0.5em{\scshape i\kern-0.25em b}\kern-0.8em\TeX}}}

\input{preamble.tex}

\usepackage[compact]{titlesec}
\titlespacing{\section}{1pt}{3ex}{2ex}
\titlespacing{\subsection}{1pt}{2ex}{1ex}
\titlespacing{\subsubsection}{1pt}{0.5ex}{0ex}

\begin{document}

\title{
\textsc{V2W-BERT}: A Framework for Effective Hierarchical Multiclass Classification of Software Vulnerabilities
}

\author{Siddhartha Shankar Das}
\affiliation{%
  \institution{Purdue University}
  \city{West Lafayette, IN}
  \country{USA}}
\email{das90@purdue.edu}
\author{Edoardo Serra}
\affiliation{%
  \institution{Boise State University}
  \city{Boise, ID}
  \country{USA}}
\email{edoardoserra@boisestate.edu}
\author{Mahantesh Halappanavar}
\affiliation{%
  \institution{Pacific Northwest National Lab}
  \city{Richland, WA}
  \country{USA}}
\email{hala@pnnl.gov}
\author{Alex Pothen}
\affiliation{%
  \institution{Purdue University}
  \city{West Lafayette, IN}
  \country{USA}}
\email{apothen@purdue.edu}
\author{Ehab Al-Shaer}
\affiliation{%
  \institution{Carnegie Mellon University}
  \city{Pittsburgh, PA}
  \country{USA}}
\email{ehab@cmu.edu}


\begin{abstract}
Weaknesses in computer systems such as faults, bugs and errors in the architecture, design or implementation of software provide vulnerabilities that can be exploited by attackers to compromise the security of a system. 
Common Weakness Enumerations (CWE) are a hierarchically designed dictionary of software weaknesses that provide a means to understand software flaws, potential impact of their exploitation, and means to mitigate these flaws. 
Common Vulnerabilities and Exposures (CVE) are brief low-level descriptions that uniquely identify vulnerabilities in a specific product or protocol.
Classifying or mapping of CVEs to CWEs provides a means to understand the impact and mitigate the vulnerabilities. 
Since manual mapping of CVEs is not a viable option, automated approaches are desirable but challenging. 

We present a novel  Transformer-based learning framework (\textsc{V2W-BERT}) in this paper. By using ideas from natural language processing, link prediction and transfer learning, our method outperforms previous approaches not only for CWE instances with abundant data to train, but also rare CWE classes with little or no data to train. 
Our approach also shows significant improvements in using historical data to predict links for future instances of CVEs, and therefore, provides a viable approach for practical applications.
Using data from MITRE and National Vulnerability Database, we achieve up to $97\%$ prediction accuracy for randomly partitioned data and up to $94\%$ prediction accuracy in temporally partitioned data.
We believe that our work will influence the design of better methods and training models, as well as applications to solve increasingly harder problems in cybersecurity. 
\end{abstract}

\maketitle
\pagestyle{plain}

\section{Introduction}
\label{sec:intro}
In order to understand and mitigate specific vulnerabilities in software products and protocols, one needs to accurately map them to hierarchically designed security dictionaries that provide insight on attack mechanisms, and thereby, means to mitigate weaknesses. Automating the mapping of vulnerabilities to weaknesses is a hard problem with significant challenges. 
In the paper, we present a novel Transformer-based framework to exploit recent developments in natural language processing, link prediction and transfer learning to accurately map vulnerabilities to hierarchically structured weaknesses, even when little or no prior information exists. 

\begin{figure}[tb]
 \centering
 \includegraphics[width=1.0\linewidth]{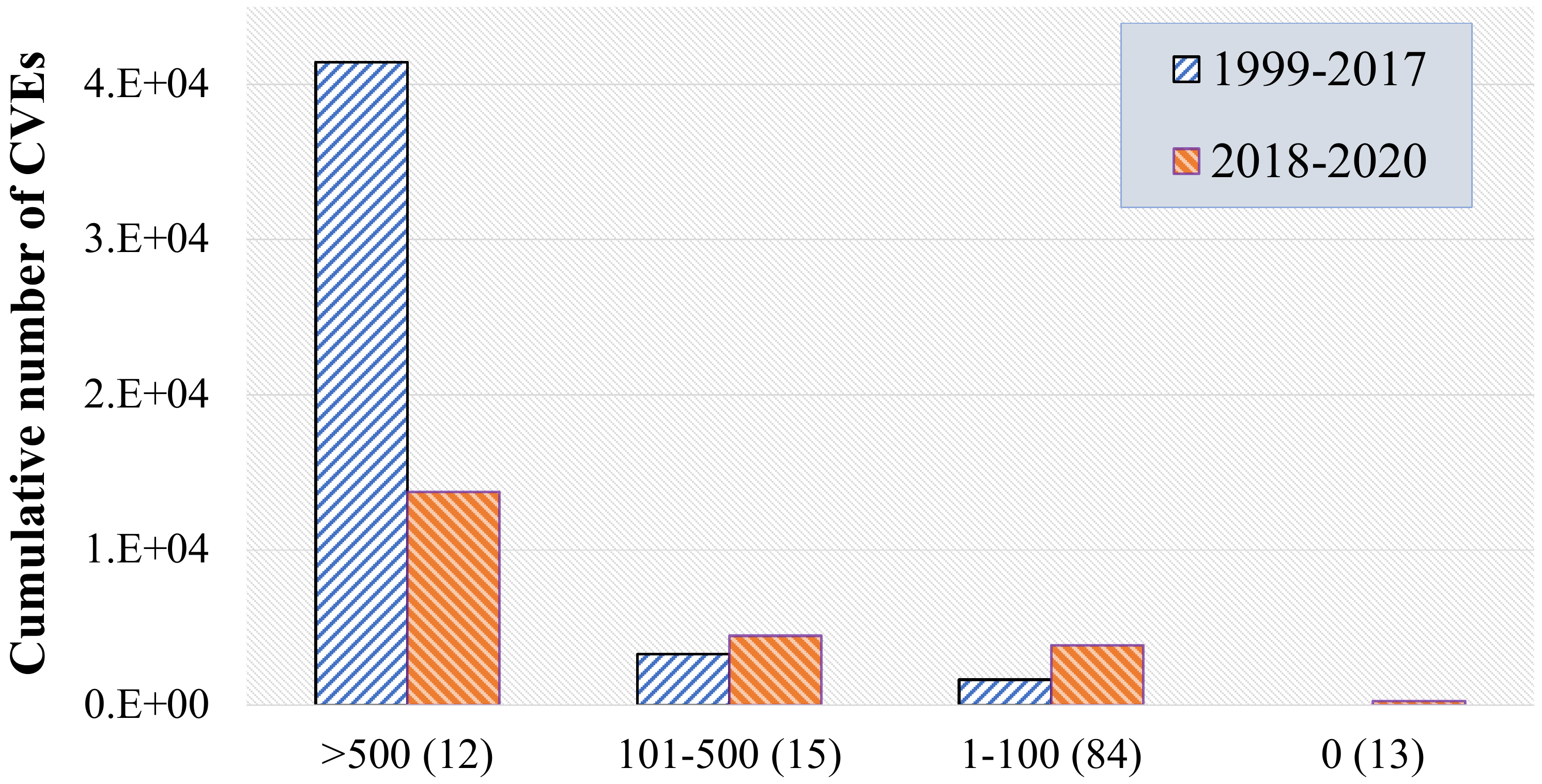}
 \caption{Distribution of the number of CVEs per CWE in the National Vulnerability Database, bucketed into four categories: 12 CWEs with  500 or more CVEs per CWE, 15 CWEs with 100 to 500 CVEs per CWE, 84 CWEs with 1 to 100 CVEs per CWE, and 13 CWEs with zero CVE. We partition the data into two time periods to simulate testing for CVEs observed in the future: 1999-2017 (used for training) and 2018-2020 (used for testing). Cumulative numbers of CVEs are plotted on the  Y-axis. The proposed framework (\textsc{V2W-BERT})  targets efficient mapping of rare instances that have not been addressed in earlier studies.}
 \label{fig:Overview}
\end{figure}
Common Weakness Enumerations (CWE)\footnote{\url{https://cwe.mitre.org}} provide a blueprint for understanding software flaws and their impacts through a hierarchically designed dictionary of software weaknesses. Weaknesses are bugs, errors and faults that occur in different aspects of software such as architecture, design, or implementation that lead to exploitable vulnerabilities. 
Non-disjoint classes of CWEs are organized in a tree structure, where higher level classes provide general definitions of weaknesses, and lower level classes inherit the characteristics of the parent classes and add further details. 
Thus, analyzing the correct path from a root to lower level nodes provides valuable insight and functional directions to learn a weakness.
For example, tracing the path from the root node, CWE-707, to a node CWE-89\footnote{\url{https://cwe.mitre.org/data/definitions/89.html}}, reveals that SQL injection (CWE-89) is caused by improper neutralization of special elements in data query logic (CWE-943), which in turn is caused by injection (CWE-74) or sent to a downstream component (CWE-707). This insight provides a means to design countermeasures even when a specific CWE node is not available~\cite{aghaei2019threatzoom}. 

In contrast, Common Vulnerabilities and Exposures (CVE)\footnote{\url{https://cve.mitre.org/cve/}} reports are uniquely identified computer security vulnerabilities, where a vulnerability is defined as a set of one or more weaknesses in a specific product or protocol that allows an attacker to exploit the behaviors or resources to compromise the system. CVEs are brief and low-level descriptions that provide a means to publicly share information on vulnerabilities.  
For example, CVE-2004-0366 provides specific description of an attack action through the execution of arbitrary SQL statement for a specific product, \texttt{libpam-pgsql} library, producing the specific consequence of SQL injection, which can then be used to compromise a system.
However, and more importantly, the CVE report does not specify the characteristics of the SQL injection that are necessary to detect and mitigate the attack~\cite{aghaei2019threatzoom}. 
This information comes from the corresponding CWE; CWE-89: SQL Injection, mentioned earlier. 

Accurate mapping of CVEs to CWEs will enable the study of the means, impact and ways to mitigate attacks;  hence it is an important problem in cyber-security~\citep{Martin2008, Jimenez2016, Cruzes2017}.
However, the problem is riddled with several challenges. A CVE can be mapped to multiple and interdependent CWEs that belong to the same path, which leads to ambiguity. CVEs are manually mapped to CWEs, which is neither scalable nor reliable. Consequently, there is a lack of high-quality mapping information. Only about 2\% of CVEs are mapped in the MITRE database. Although NVD provides a higher percentage of mapping, about 71\%, the number of CWEs that are mapped is considerably small (about 32\%). As of February 2021, there are a total of $157,325$ CVEs registered in the NIST National Vulnerability Database (NVD), and $916$ CWEs in the MITRE CWE database. Since new CVEs are created at a fast pace, manual mapping of CVEs is not a viable approach. 
Therefore efficient methods to automate the mapping of CVEs to CWEs are critical to address the ever increasing cybersecurity threats. We propose a novel method in this paper to address this challenging problem.

Automated mapping is limited by several challenges such as lack of sufficient training data, semantic gaps in the language of CVEs and CWEs, and non-disjoint hierarchy of CWEs classes. Our work focuses on one of the hardest problems in mapping CVEs --- rare CWE classes that do not have any CVEs mapped to them. As illustrated in Figure~\ref{fig:Overview}, a significant number of CVEs are currently mapped to a small set of CWE classes. 
Currently, about {\bf 70\%} of the CWE classes have fewer than 100 CVEs for training,
about 10\% have no CVEs mapped to them, and 
only 10\% have more than 500 CVEs. 
The current approaches of classification work well only when a sufficient amount of data is available to train~\citep{na2016study, aota2020automation, neuhaus2010security, rehman2012software}. Although recent efforts using neural networks and word embedding based methods to process CVE reports have showed better performance~\citep{aghaei2019threatzoom, han2017learning,nakagawa2019character}, they fail when little or no training data exists. Consequently, a large set of rare CWEs are completely ignored in literature. 
A second challenge that we address in this work is the practical scenario of classifying the vulnerabilities based on past data $(1999-2017)$ to predict future data $(2018-2020)$. Furthermore, rare CWE cases have been appearing more frequently in recent years,  thus making the task even harder. 

In this paper, we present a novel Transformer-based~\citep{vaswani2017attention} learning framework, \textsc{V2W-BERT}, that outperforms existing approaches for mapping CVEs to the CWE hierarchy at finer granularities. 
In particular, \textsc{V2W-BERT} is especially effective for rare instances.  
The Bidirectional Encoder Representations from Transformers (BERT) is designed to pre-train deep bidirectional representations from unlabeled text by jointly conditioning on both the left and right sides of the context of a text token  during the training phase~\citep{devlin2018bert}. 
BERT is trained on a large text corpus, learning a deeper and intimate understanding of how language works, which is useful for downstream language processing tasks. Pre-trained BERT models can be enhanced with additional custom layers to customize for a wide range of Natural Language Processing (NLP) tasks~\citep{devlin2018bert,sun2019fine}. 
We exploit this feature to transfer knowledge to the security domain and use it for mapping CVEs.

The second aspect of novelty in our work comes from the formulation of the problems as a {link prediction} problem that is different from previous formulation. In particular, we use the Siamese model~\citep{chicco2020siamese} to embed  semantically different text forms in CVEs and CWEs into the same space for mapping through link prediction --- associate the best link from a CVE to a CWE. 

\noindent
\textbf{Contributions:} The key contributions of our work are as follows:
\begin{enumerate}[leftmargin=*,noitemsep,topsep=1pt]
    \item We present a novel Transformer-based learning framework, \textsc{V2W-BERT}, to classify CVEs into CWEs (\S\ref{sec:method}), including a detailed ablation study (\S\ref{sec:ablation}). 
    Our framework exploits both labeled and unlabeled CVEs, and uses pre-trained BERT models in a Siamese~\citep{chicco2020siamese} architecture to predict links between CVEs and CWEs (\S\ref{subsec:pretrain}). Rare and unseen vulnerabilities are classified using a transfer learning procedure. 
    The Reconstruction Decoder (\S\ref{ssec:RD}) ensures that the  pre-trained model on unsupervised data is not compromised by overfitting the link prediction task.
    
    
    \item This is the first work to formalize the problem of mapping multiple-weakness definitions into a single vulnerability (multi-class) as a link prediction task. 
    The hierarchical relationships among the CWEs are also incorporated during training and prediction phases.
    Unlike the traditional classification based methods, adding new CWE definitions to the corpus does not require changes to the model architecture.
    
    \item \textsc{V2W-BERT} outperforms related approaches for all types of CWEs (both rare and frequently occurring) (\S\ref{sec:comparison}). 
    We simulate a  challenging real-world scenario in our experiments where future mappings (2018-2020) are predicted based on past years' (1999-2017) data. 
    We predict the CWEs of a vulnerability to finer-granularities (root to the leaf node), and the user can control the precision.
    
    \item For frequently occurring cases, \textsc{V2W-BERT} predicts immediate future (2018) mappings with \textbf{89\%-98\%} accuracy for precise and relaxed prediction (definitions of these modes of prediction are provided in \S\ref{sec:experimentalresults}). For rarely occurring CVEs, the proposed method achieves \textbf{48\%-76\%} prediction accuracy, which is 10\% to 15\% higher than the existing approach. 
    Additionally, the proposed method can classify completely unseen types of CWEs with up to \textbf{61\%} accuracy. We believe that this feature enables us to detect if and when a new CWE definition becomes necessary.
\end{enumerate}
\vspace{0.2cm}

To the best of our knowledge, this is the first work to propose a novel Transformer-based framework that builds on link prediction to efficiently map CVEs to hierarchically-structured CWE descriptions. The framework not only performs well for CWE classes with abundant data, but also for rare CWE classes with little or no data to train, along with the power to 
map as yet unseen CVEs to existing or new CWEs.
Therefore, we believe that our work will motivate the development of new methods as well as practical applications of the framework to solve increasingly challenging problems in automated organization of shared cyber-threat intelligence~\cite{WAGNER2019101589}.

\section{Preliminaries \& Related Work}
\label{sec:background}

\subsection{Problem formulation}
\label{subsec:formulation}
The Common Vulnerabilities and Exposures (CVEs) reports comprise the input text data, and the Common Weakness Enumerations (CWEs) are the target classes. The CWEs have textual details (Name, Description, Extended Description, Consequences, etc.), which are 
ignored in classification based methods. To utilize CWE descriptions and make the model flexible, we convert this multi-class multi-label problem into a binary link prediction problem. We propose a function, $F_\theta$, that takes a CVE-CWE description pair $(u_{cve},~v_{cwe})$ and returns a confidence value measuring their association:
\begin{equation}
    l = F_\theta(u_{cve},v_{cwe}).
\label{eqn:1}
\end{equation}

Here, $F_\theta$ is a learnable function and the vector $\theta$ denotes learnable parameters. If a particular CVE ($u_{cve}$) is associated to a CWE ($v_{cwe})$, then the function $F_\theta$ returns a value $l\approx 1$; and, $l\approx 0$ otherwise.

To learn $\theta$, both positive and negative links from the known associations are used. If a CVE has a known mapping to some CWE in the hierarchy, we consider all associations between them and their ancestors as positive links. The rest of the CVE-CWE associations are negative links. To predict the CWEs to be associated with a CVE report, we find the link with the highest confidence value in the hierarchy,  from the root to a leaf node, using $F_\theta$. 
The function $F_\theta$ also helps to easily incorporate new CWE definitions into the classification model.

\subsection{Brief Overview of BERT}
\label{subsec:bertpreliminaries}
BERT~\citep{devlin2018bert} stands for Bidirectional Encoder Representations from Transformers.
Transformers are attention-based Neural Networks that can effectively handle sequential data like texts by learning the relevance to the far away tokens concerning the current token~\citep{vaswani2017attention}.
Unlike directional models, which read the text input sequentially (left-to-right or right-to-left),  BERT is a bidirectional model that learns the context of a word based on its surroundings.
Training on large unlabeled text corpus helps BERT learn how the underlying languages work.
Devlin et al.~\citep{devlin2018bert} reported two BERT models, BERT\textsubscript{BASE} ($L=12, H=768, A=12$, Total parameters=110M), BERT\textsubscript{LARGE} ($L=24, H=1024, A=16$, Total parameters=340M) where $L, H, A$ stand for number of layers (Transformer blocks), hidden size, and number of self-attention heads,  respectively. 

The original BERT models are pre-trained considering two tasks: ($i$) Masked Language Model (LM), and ($ii$) Next Sentence Prediction (NSP). In the Masked LM task, $15\%$ of random tokens are masked in each text sequence. Among those masked tokens, $80\%$ are replaced with token \texttt{[MASK]}, $10\%$ are replaced with random tokens, and $10\%$ are kept the same. These masked inputs are fed through the BERT encoder model, and the hidden states are passed to a decoder containing a linear transformation layer with \textit{softmax} activation over the vocabulary. The model is optimized using cross entropy loss.

As for Next Sentence Prediction (NSP) task, a pre-training batch consists of pairs of sentences $C, D$ where $50\%$ of the time $D$,  the  sentence next to $C$,  appears in the training samples, and for the remainder  they do not. NSP helps downstream Question Answering (QA) and Natural Language Inference (NLI) tasks by directly learning the relationship between sentences. 
The pre-trained BERT models (BERT\textsubscript{BASE},  BERT\textsubscript{LARGE}) are trained over BooksCorpus ($800M$ words) and the English Wikipedia ($2500M$ words) dataset, considering both MLM and NSP tasks together. 

BERT\textsubscript{BASE} uses WordPiece embeddings with 30,522 vocabulary tokens to convert text sequences to vector forms. The first token is always \texttt{[CLS]} and end of a sentence is represented with \texttt{[SEP]}. The final hidden state corresponding to this \texttt{[CLS]} token usually represent the whole sequence as an aggregated representation. 
In this work, BERT\textsubscript{BASE} is used, and other variants of sequence representation are considered through different pooling operations. 

\subsection{Related Work}

Several studies have investigated the CVE to CWE classification problem. However, \textsc{V2W-BERT} is the first approach that formulates the problem as a link prediction problem using Transformers. 
Recent work by Aota et al.~\citep{aota2020automation} uses Random Forest and a new feature selection based method to classify CVEs to CWEs. This work only uses the $19$ most frequent CWE definitions  and ignores CWEs with fewer than $100$ instances. It achieves $F1_{macro}$-Score of $92.93\%$ for classification. Further, it does not support multi-label classification and does not consider the hierarchical relationships within CWEs. All these limitations are addressed in our work. 

Na et al.~\citep{na2016study} predict CWEs from CVE descriptions using a Na\"ive Bayes classifier. They focused only on the most frequent 2-10 CWEs without considering the hierarchy. When the number of CWEs considered increases from 2 to 10, their accuracy drops from $99.8\%$ to $75.5\%$.
Rahman et al.~\citep{rehman2012software} use Term Frequency-Inverse Document Frequency (TF-IDF) based feature vector and Support Vector Machine (SVM) technique to map CVEs to CWEs. They use only 6 CWE classes and 427 CVEs without considering hierarchy. 

Recent work by Aghaei et al.~\citep{aghaei2019threatzoom} uses TF-IDF weights of the vulnerabilities to initialize single layer Neural Networks (NNs). They use CWE hierarchy to predict classes iteratively. However, this is a shallow NN with only one layer, and comparative performance with more complex networks is not discussed in their work. Further, they consider all classes with scores higher than a given threshold as a prediction. This approach decreases the precision of prediction and is less desirable when precise predictions are needed, a limitation that is addressed in our work. 
Depending on the level of hierarchy, they achieve $92\%$ and $94\%$ accuracy for a random partition of the dataset. In contrast, we study a more representative partition of data based on time.

We note that each study uses different sets of CVEs for learning and testing. The choice of the number of  CWEs used and evaluation methods are also different. Therefore, there is no consistent way to compare the accuracy numbers presented by different authors. Some studies use CVE descriptions to perform fundamentally different tasks than mapping to CWEs. For example, Han et el.\citep{han2017learning} and Nakagawa et al.~\citep{nakagawa2019character} use \texttt{word2vec} for word embedding and Convolutional Neural Network (CNN) to predict the severity of a vulnerability (score from 0 to 10). Neuhaus et al. \citep{neuhaus2010security} use Latent Dirichlet Allocation (LDA) to analyze the CVE descriptions and assign reports on 28 topics.


To the best of our knowledge, \textsc{V2W-BERT} is the first BERT~\citep{devlin2018bert} based method to classify CVEs into CWEs. We fine-tune the pre-trained BERT model with CVE and CWE descriptions, and then learn $F_\theta$ (Equation~\ref{eqn:1}), using a Siamese network of BERT. A Siamese network shares weights while working in tandem on two different inputs to compute comparable outputs.
A few recent studies have used the Siamese BERT architecture for information retrieval and sentence embedding tasks~\citep{reimers2019sentence, lu2020twinbert}.
Reimers et al.~\citep{reimers2019sentence} proposed Sentence-BERT (SBERT), which uses Siamese and triplet network for sentence pair regression and achieves the state-of-the-art performance in Semantic Textual Similarity (STS)~\citep{agirre2015semeval}.
\textsc{V2W-BERT} is conceptually similar to SBERT, but with notable differences. \textsc{V2W-BERT} has a different architecture where Reconstruction Decoder is coupled with the Siamese network to preserve context to improve performance in classifying rare and unseen vulnerabilities. Further, \textsc{V2W-BERT} is designed to classify CVEs into CWEs hierarchically, and therefore, has significantly different training and optimization processes.
\begin{figure*}[!hbtp]
\centering
 \resizebox{0.7\linewidth}{!}{

    \centering
    \tikzset{
    block/.style={
      draw, 
      rectangle, 
      thick,
      minimum height=0.5cm, 
      minimum width=3cm, align=center
      },
     roundblock/.style={
      draw, 
      rectangle, 
      rounded corners,
      thick,
      minimum height=0.5cm, 
      minimum width=3cm, align=center
      },
      component/.style={
      thick,
      minimum height=0.5cm, 
      minimum width=3cm, align=center
      },
    line/.style={->,>=latex'},
    vhilit/.style={draw=black, thick, dotted,
    inner sep=1em,
    },
    hhilit/.style={draw=black, thick, densely dotted,
        inner xsep=0.5em,
        inner ysep=.5em},
    outblock/.style={draw=black, thick,
        inner xsep=0.5em,
        inner ysep=.5em}
    }
    
    \begin{tikzpicture}[node distance=2.0cm]
    

    \node (cwedescription) {CWE description, $v_{cwe}$};
    \node [left =1.0cm of cwedescription]  (cvedescription) {CVE description, $u_{cve}$};
    
    \node[block, above =0.5cm of cvedescription] (bertleftfixed) {BERT, \\layer $1$ to $f$, Fixed};
    
    \node[block, above =0.2cm of bertleftfixed] (bertlefttrainable) {BERT, \\layer $f$ to $A$, Trainable};
    
    \node[outblock,fit=(bertleftfixed) (bertlefttrainable)](bertleft) {};

    \node[block, above =2.0cm of bertlefttrainable] (bertleftpool) {Pooling, $\vx_{cve}$};

    \node[block, above =0.5cm of cwedescription] (bertrightfixed) {BERT, \\layer $1$ to $f$, Fixed};
    
    \node[block, above =0.2cm of bertrightfixed] (bertrighttrainable) {BERT, \\layer $f$ to $A$, Trainable};
    
    \node[outblock,fit=(bertrightfixed) (bertrighttrainable)](bertright) {};
    
    \node[block, above =2.0cm of bertrighttrainable] (bertrightpool) {Pooling, $\vx_{cwe}$};
    
    \node[block, above of=bertleftpool, right of=bertleftpool] (combine) {Combine, $\vx_{cve},\vy_{cwe}$};
    
    \node[block, above =0.5cm of combine] (linkclass) {Link Classifier};
    
    \node[roundblock, above =0.5cm of linkclass] (linkclassloss) {Classification Loss};
    
    \node[component, above =0.5cm of linkclassloss] (lp) {Link Prediction (LP)};
    
    \draw[line] (cvedescription.north) -- (bertleft.south);
    \draw[line] (cwedescription.north) -- (bertright.south);
    \draw[line] (bertleft.north) -- (bertleftpool.south) node [midway,fill=white]{Hidden State, $(T, H)$};
    \draw[line] (bertright.north) -- (bertrightpool.south) node [midway,fill=white]{Hidden State, $(T, H)$};
    
    \draw[line] (bertleftpool.north) -- (combine.south) node [midway,fill=white] {$\vx_{cve}$};
    \draw[line] (bertrightpool.north) -- (combine.south) node [midway,fill=white] {$\vy_{cwe}$};
    \draw[line] (combine.north) -- (linkclass.south);
    \draw[line] (linkclass.north) -- (linkclassloss.south);
    
    \node[hhilit,fit=(bertleft) (bertleftpool) (bertright) (bertrightpool) (combine) (linkclass) (linkclassloss)] {};

    \node[block, left =0.9cm of bertleftpool] (leftlm) {Masked \\Language Model (LM)};
    
    \node[roundblock, above of=leftlm] (leftlmross) {CVE LM Loss};
    
    \node[component, above =0.5cm of leftlmross] (lrd) {Reconstruction Decoder (RD)};
    
    \draw[line] (bertlefttrainable.west) --  (leftlm.south)  node [midway,fill=white]{Hidden State, $(T, H)$};
    \draw[line] (leftlm.north) -- (leftlmross.south);
    
    \node[hhilit,fit= (leftlm) (leftlmross)] {};

    \node[block, right =0.9cm of bertrightpool] (rightlm) {Masked \\Language Model (LM)};
    
    \node[roundblock, above of=rightlm] (rightlmross) {CWE LM Loss};
    
    \node[component, above =0.5cm of rightlmross] (rrd) {Reconstruction Decoder (RD)};
    
    \draw[line] (bertrighttrainable.east) -- (rightlm.south)  node [midway,fill=white]{Hidden State, $(T, H)$};
    \draw[line] (rightlm.north) -- (rightlmross.south);

    \node[hhilit,fit=(rightlm) (rightlmross)] {};
    

\end{tikzpicture}
}
\caption{An overview of the architecture of \textsc{V2W-BERT} framework with the Link Prediction module (shown in the middle) and the Reconstruction Decoder modules (shown on the left and right).}
\label{fig:overallmodel}
\end{figure*}
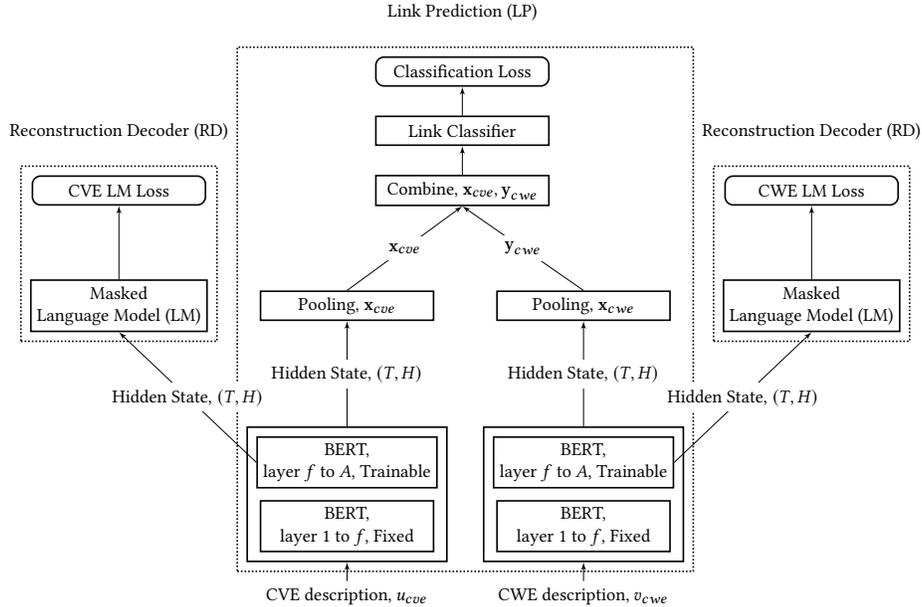
\vspace{-0.3cm}

\section{A Novel Framework: \textsc{V2W-BERT}}
\label{sec:method}

In this section, we present a novel framework \textsc{V2W-BERT} to classify CVEs to CWEs hierarchically. \textsc{V2W-BERT} optimizes the learnable parameters $\theta$ of $F_\theta$ (\S\ref{subsec:formulation}) in two steps. In the first step,  the pre-trained BERT language model is further fine-tuned with CVE/CWE descriptions specific to cyber security. 
In the second step, the trained BERT model is employed in a Siamese network architecture to establish links between CVEs and CWEs. The architecture takes a specific CVE-CWE pair as input, and predicts whether the CVE belongs to the CWE or not,  with a  confidence value. \textsc{V2W-BERT} includes a Mask Language Model (LM) based Reconstruction Decoder to ensure that the descriptions' contexts are not changed too much during the training process.

Figure~\ref{fig:overallmodel} shows the overall architecture of the \textsc{V2W-BERT} framework. \textsc{V2W-BERT} contains two primary components: ($i$) Link Prediction (LP), and ($ii$) Reconstruction Decoder (RD). 
The LP module's primary purpose is to map CVEs with CWEs while the RD module preserves the context of the descriptions of CVEs and CWEs.  During the backpropagation step, the trainable BERT layers are updated while optimizing LP and RD loss simultaneously. 
Figure~\ref{fig:overallmodel} shows a simplified architecture where 
the attention, fully connected, dropout, and layer-normalization layers have been omitted.

\subsection{Unsupervised Pre-training of BERT}
\label{subsec:pretrain}
Specific downstream inference tasks benefit from pre-training BERT with language associated with the domain-specific unlabeled data and the addition of custom Neural Network layers to the base model.
To incorporate the cyber-security specific data on top of the base model, we  pre-train BERT further with  CVE and CWE descriptions. This is useful as a significant amount of CVE descriptions 
are not labeled and thus do not help with  supervised learning.
Since the pre-training process does not require CWE class labels, we utilize both labeled and unlabeled CVE descriptions to learn the cyber-security context.
The original BERT model is trained considering Masked Language Model (LM) and Next Sentence Prediction (NSP) tasks.  Like NSP, CVE and CWE are linked using the Link Prediction (LP) component as the second step of the \textsc{V2W-BERT} algorithm. Therefore, the BERT encoder is tuned on the Masked LM task only over available CVE and CWE descriptions. 
All layers of BERT are allowed to be  updated in the pre-training step incorporating the cyber-security context. 
Section~\ref{subsec:pretrainarchitecture} in the Appendix shows the architecture of the Masked Language Model in more detail.

\subsection{Link Prediction Component}
In the original problem, $l=F_\theta (u_{cve}, v_{cwe})$, both CVE and CWE descriptions  need to be processed together to establish links between them. 
There are many ways to tackle this. For example,  TF-IDF or word embeddings (word2vec, glove, etc.) could be used to get vector representations of CVEs and CWEs, and these representations could be combined and classified with any learnable method that returns confidence about the association. However, the pre-trained BERT model knows the context of this problem domain, and can map relevant descriptions to similar vector spaces better than word embeddings~\citep{reimers2019sentence}. Furthermore, we need BERT to be tuned for the function $F_\theta$, and the multi-layer Neural Network is the most compatible classification approach.

Therefore, in the Link Prediction (LP) component of \textsc{V2W-BERT}, the pre-trained BERT model is used to transform the CVE/CWE description.
We fix the parameters of first $f$ out of $A$ layers ($A=12$ in BERT\textsubscript{BASE}) to allow minimal changes to the model to preserve previously learned context~\citep{sun2019fine}. 
We used $f=9$ in this study. 
LP adds a pooling layer on top of the pre-trained BERT encoder model to get a vector representation of the input sequence. These individual representations are then combined and passed through a classification layer with the \textit{softmax} activation function. The output values create the relationship between a CVE and a CWE description with a degree of confidence. 

\textbf{Pooling:} By default, the hidden state corresponding to the  \texttt{[CLS]} token from the BERT encoder is considered as a pooled vector representation.
However, recent work~\citep{sun2019fine} has shown that other pooling operations can perform better depending on the problem. Two additional pooling methods \texttt{MAX}-pooling 
(it takes \texttt{MAX}  of the representation vectors of all tokens),
and \texttt{MEAN}-pooling (which takes the \texttt{MEAN} of the vectors)  are considered in ouor work. 
The pooled representations are passed through another transformation layer to get the final vector representation. 
In the CVE classification task, we found \texttt{MEAN}-pooling to be the best performing. The pooled vector representations are denoted as $\vx_{cve}$ for a CVE and$~\vy_{cwe}$ for a CWE. 

\textbf{Combination:}  The pooled representations of input sequence pair can be combined in different ways~\citep{cer2018universal,reimers2019sentence}. Some common operations are:  Concatenation, multiplication, addition, set-operations, or combinations of these. In the current problem, concatenation of absolute difference and multiplication $(|\vx_{cve}- \vy_{cwe}|,\vx_{cve}\times \vy_{cwe})$ operation has shown best performance.
Appendix~\ref{subsec:combination} shows that there are  significant differences in the results from these choices. 

\textbf{Link Classification:} The combined representations are classified into the link and unlink confidence values using the linear output layer with two neurons and \textit{softmax} activation function. The \textit{softmax} value ranges between $[0,1]$ and represents the confidence value of associating a CVE to a CWE. 

For a specific CVE-CWE pair, if the link value is higher than the unlink value, then the CVE is associated with that CWE. A single neuron can also classify a link/unlink when the value is  close to $1.0$, indicating a high link association. However, experiments  show that an output layer with two neurons outperforms a single neuron classifier. The cross-entropy loss is used to optimize link prediction: 
\begin{equation}
    CL(u_{cve}, v_{cwe})=CE(LP_\theta(u_{cve}, v_{cwe}), REAL(u_{cve}, v_{cwe})),
    \label{eq:clloss}
\end{equation}
where, $CL(u_{cve}, v_{cwe})$ is the link classification loss between predicted and real values of the CVE-CWE relation. 
$LP_\theta(u_{cve},v_{cwe})$ generates a $2$-dimensional vector where first and second indices represent unlink and link association confidence values, respectively.
If $u_{cve}$ belongs to $v_{cwe}$, ideally these values should be  $\approx 0$ for first index, and $\approx 1$ for the second index.


\subsection{Reconstruction Decoder Component}
\label{ssec:RD}
The classification challenge comes from three types of CVEs associated with rare CWEs classes: 
($i$) The CVEs belonging to a CWE class with few training instances, 
($ii$) the CVEs of a particular CWE that appear in the test set but not in the training set, 
and ($iii$) CVEs  with description styles that differ from  the training  set,  or instances where the labels are erroneous. 

The advantage of transfer learning is that it helps classify cases with few training instances~\citep{sun2019fine} as pre-trained BERT can produce correlated transformed vector representations from similar input sequences. The Link Prediction (LP) component learns to relate a CVE with the available CWEs by establishing links even when the  training instances are few or do not exist.  

For a new  CVE type, we expect to have a low link association value with CWEs that exist in the training set (due to negative training links), and a high value for CWEs not included in the training set with similar text descriptions. 
However, due to learning bias towards available CWEs in Link Prediction (LP), we will have a higher link association to existing CWEs compared to new CWEs. 
Therefore, if we could preserve the original context that BERT learned during the pre-training phase while changing the LP model, it could improve the performance for rare CVE cases,  and  for completely unseen CWE classes. Note that for unseen cases this approach would work only if the corresponding CVE and CWE descriptions have some textual similarity. 
Preserving context can also be useful for detecting unusual or differently styled CVE descriptions during the test as they may not create any links with the available CWEs.

To preserve context while updating LP, we add a Reconstruction Decoder (RD) component (Figure~\ref{fig:overallmodel}). When the BERT encoder transforms a CVE/CWE description, the last hidden state is passed to the Masked Language Model (LM) and optimized for Masked tokens. LP and RD share BERTs' hidden states, and the trainable layers are updated considering both link classification loss and reconstruction loss simultaneously. In this way, \textsc{V2W-BERT} trains for link classification while preserving context. Cross-Entropy loss is used to optimize the difference between original input and reconstructed tokens.

Let $RL(u_{cve})$ denote the reconstruction loss of an input sequence $u_{cve}$;  and $RD_\theta(BERT_A(u_{cve}))$ be a reconstruction decoder that takes the last hidden state of BERT and reconstructs masked $u_{cve}$ tokens. We can express the reconstruction loss as follows:
\begin{align}
    \begin{array}{l}
    RL(u_{cve}) = CE(RD_\theta(BERT_A(u_{cve})), MASKED(u_{cve})),\\
    RL(v_{cwe}) = CE(RD_\theta(BERT_A(v_{cwe})), MASKED(v_{cwe})).
  \end{array}
  \label{eq:rlloss}
\end{align}
\subsection{Training Details}
To learn the parameters  $\theta$ of the model $F_\theta$, we have to train \textsc{V2W-BERT} with positive and negative link mappings between CVEs and CWEs. A single CVE can belong to multiple CWEs at different levels of the hierarchy. According to the MITRE classification,  a CWE can have multiple parents and multiple children. When a CVE belongs to a CWE, that CVE-CWE pair is considered a positive link, and all ancestor CWEs of that weakness are also considered as  positive links. The remaining CWEs available during training are used for negative links ( unlinks). 

Let $B_{cve}$ be a mini-batch of CVEs selected randomly. The set $\mathit{CWE}(u)$ denotes the  CWEs associated with a vulnerability $u$, and $ancestors(w)$ is the set of all ancestors of the weakness $w$. Similarly, $U_{cwe}$ is a set of CWEs available only to the training data.
The positive and negative links ($P, N$) for training are generated as follows:
\begin{align}
    P&= \bigcup_{u \in B_{cve}} \bigcup_{w \in \mathit{CWE}(u)} \{(u,w)\cup\{(u,v): v \in ancestors(w)\}\};\\
    N&= \bigcup_{u \in B_{cve}} \{(u,w):w \in \text{$k$ randomly in}\{U_{cwe}-\mathit{CWE}(u)\} \}.
\end{align}


Using Equations~\ref{eq:clloss} and~\ref{eq:rlloss}, the losses for the $(P, N)$ links from the LP and RD components can be expressed as, 
\begin{align}
    L_P &=\sum_{(u_{cve},v_{cwe})\in P} CL(u_{cve}, v_{cwe})+ RL(u_{cve})+RL(v_{cwe});\\
    L_N &=\sum_{(u_{cve},v_{cwe})\in N} CL(u_{cve}, v_{cwe})+ RL(u_{cve})+RL(v_{cwe}).
\end{align}
Here, $CL$ and $RL$ refer to link classification and reconstruction loss, respectively.
Since a single CVE can belong only to a few CWEs,  only a few positive link pairs are present in a batch compared to the possible negative links. In the loss function, it is necessary to balance $P$ and $N$ to prevent bias, and this can be prevented either by repeating positive links in a batch or putting more weight on positive links $P$.
The total loss,  $L_{B_{cve}}$, in a mini-batch of CVEs is given by:
\begin{align}
    L_{B_{cve}} &= \gamma_1\times L_P+\gamma_2\times L_N. 
\end{align}
The parameters $\theta$ of the model $F_\theta$ are updated after processing the links from each mini-batch.

\subsection{CVE to CWE Prediction using \textsc{V2W-BERT}}
\textsc{V2W-BERT} considers the same CWE hierarchy during learning and prediction. CVE data in NVD use only a subset of the CWEs from MITRE, and the hierarchical CWE relations available in NVD omit some of the parent-child relations available in MITRE. Therefore, we use the same 124 CWEs used in NVD, but their hierarchical relationships are enriched using the data from  MITRE\footnote{\href{https://drive.google.com/file/d/1YHQJOPvWxfxpxy1NyjQ35v5nnBNEEpdg/view?usp=sharing}{Partial CWE hierarchy extracted from MITRE}}.

These 124 CWEs are distributed in three levels in the hierarchy, with 34 in the first level, 78 in the second level, and 16 in the third level. Some CWEs have multiple parents in different levels and are counted twice. 
At the first level, there are 34 CWEs, and the prediction is made among these 34 CWEs initially. For a single CVE, we create 34 CVE-CWE pairs and get the predicted link values from the Link Prediction (LP) component. The link value with the highest confidence is considered as the CWE prediction. Next, we consider the children of the predicted CWE,  and continue until we reach a leaf node.

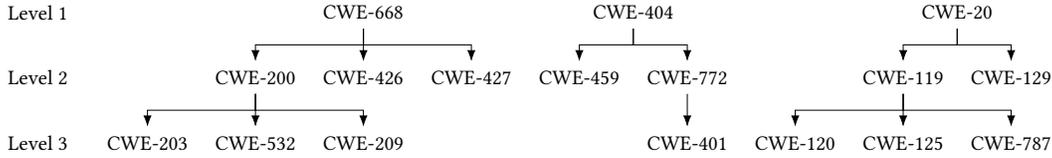
\begin{figure*}
    \centering
    \resizebox{0.8\linewidth}{!}{
    
    \tikzset{%
    every node/.style={%
        outer sep=0,
        minimum size=0.5cm,
        minimum height=5mm,
        align=center,
        anchor=north,
    },
    Trt/.style={%
    },
    Comment/.style={%
        draw=none,
        inner sep=0mm,
        outer sep=0mm,
        minimum height=5mm,
        align=right
    },
}

\forestset{myst/.style={%
    for tree={%
        parent anchor=south, 
        child anchor=north,
        l sep=0.5cm,
        edge path={\noexpand\path[\forestoption{edge},-{Latex}] 
         (!u.parent anchor) |- ($(!u.parent anchor)!.5!(.child anchor)$) -| (.child anchor)
         \forestoption{edge label};}
        }
    }
}
    
    \begin{forest} myst,
[,phantom
    [CWE-668, Trt, name=lvl3,
        [CWE-200, Trt, name=lvl2,
            [CWE-203, Trt, name=lvl1]
            [CWE-532, Trt]
            [CWE-209, Trt]
        ]
        [CWE-426, Trt]
        [CWE-427, Trt]
    ]
    [CWE-404, Trt,
        [CWE-459, Trt]
        [CWE-772, Trt,
            [CWE-401, Trt]
        ]
    ]
    [CWE-20, Trt,
        [CWE-119, Trt,
            [CWE-120, Trt]
            [CWE-125, Trt]
            [CWE-787, Trt]
        ]
        [CWE-129, Trt]
    ]
]
\node[left=0.5cm of lvl1, Comment](l1) {Level 3};
\node[above=0.5cm of l1, Comment](l2) {Level 2};
\node[above=0.5cm of l2, Comment](l3) {Level 1};
\end{forest}
    }
\caption{Partial hierarchy of CWE extracted from MITRE to demonstrate how precise and relaxed prediction is performed.}
\label{fig:predictionhierarchy}
\end{figure*}

To illustrate, Figure~\ref{fig:predictionhierarchy} shows a partial hierarchy of CWEs extracted from MITRE. At the first level, there are three CWEs (`CWE-668`, `CWE-404`, `CWE-20`), and prediction will be made among these three at first. If `CWE-668` is predicted, we predict the next weakness among its three children (`CWE-200`, `CWE-426`, `CWE-427`),  and continue until it reaches a leaf node.

Based on the user preference it is useful to have precise or relaxed prediction. For a precise prediction, we can select the best ($k_1=1$) from first level, the best ($k_2=1$) from second level (if exists), 
and the best ($k_3=1$) from the third level (if exists). 
For a relaxed prediction, we can select the top $k_1\leq 5$  confident CWEs from the first level, the top $k_2\leq 2$ from each of their children in the second level, and the best $k_3\leq 2$ from the third level. This type of user-controlled precision is useful to get better confidence about the predictions.

\section{Experimental Results}
\label{sec:experimentalresults}
We begin by discussing  experimental settings for CVE to CWE classification, and then  in an ablation study, we evaluate each component of the \textsc{V2W-BERT} framework  to investigate how 
the best performance may be obtained. Finally, we compare the \textsc{V2W-BERT} framework  with related approaches.

\subsection{Experimental Settings}
\noindent\textbf{Dataset Description}\\
The Common Vulnerabilities and Exposure (CVE) dataset is collected from the NVD\footnote{https://nvd.nist.gov/} website. After processing and filtering, we get 
$137,101$ usable CVE entries dating from 1999 to 2020 (May). Among these $82,382$ CVE entries are classified into CWEs. 
MITRE
categorizes CWEs based on Software Development, Hardware Design, and Research Concepts. Research Concepts cover all of the $916$ weaknesses, but NVD uses only $124$ of these 
CWEs. We use the same 124 CWEs used in NVD, but also include their hierarchical relations from MITRE. 

We simulate real-world CVE-CWE classification scenarios by temporally partitioning  the dataset by years. CVEs from the year 1999-2017 are included in the training set, CVEs of the year 2018 are used as Test Set 1, and CVEs of 2019-2020 are used as Test Set 2. Test Set 1 and Test Set 2 act as a near-future and far-future test cases, respectively.  There are  $46,003$ instances in training, $14,176$ instances in Test Set 1, and $22,203$ instances in  Test Set 2. 
This  temporal split creates a forecasting scenario when future 
CVEs need to be classified using currently available data, but it makes accurate  CVE classification more difficult as  CVE description styles change with time, and new CVEs occur in more recent years.
We also report results from a  random partition of the data (stratified k-fold cross-validation),  where we randomly take $70\%$ of the data from each category
for training, $10\%$ for validation of early stopping criteria and for hyperparameter settings, and $20\%$ for testing.
\newline\newline
\noindent\textbf{\textsc{V2W-BERT} Settings}\\
In the pre-training phase of \textsc{V2W-BERT}, we allow weights of all BERT\footnote{\href{https://huggingface.co/bert-base-uncased}{https://huggingface.co/bert-base-uncased}} layers to be updated. The model is trained for 25 epochs with a mini-batch size of 32.
In the CVE to CWE association phase, we freeze the first nine out of twelve layers of BERT and allow the last three layers to be updated. 
The model is trained for 20 epochs with a mini-batch size of 32. 
The number of random negative links for a CVE is set to $32$, and positive links are repeated (or can be weighted) to match the number of negative links  to prevent bias. The Adamw~\citep{loshchilov2017decoupled} optimizer is used with a learning rate of $2e^{-5}$,  and with warm-up steps of 10\% of the total training instances.
For training the \textsc{V2W-BERT} algorithm,  we used two Tesla P100-PCIE-16GB GPUs and 20 CPUs. \textsc{V2W-BERT} processes about $5K$ links for a mini-batch of 32 CVEs. For optimization, we compute the pooled representation of the CVE and CWE mini-batches separately, and combine them later as per training links ($P$, $N$).
For each configuration, the experiments were repeated   five times and the results were averaged. 
The method with the best performance is highlighted in bold in the Tables.
\newline\newline
\noindent\textbf{Evaluation Process}\\
The 124 CWEs are distributed in three levels in the MITRE hierarchy, and the CWEs that each CVE belongs to are  predicted at each level down the hierarchy. 
There are $34$  first-level CWEs, and each class has three child CWEs on an average, with  a maximum of nine. At the second level, each CWE has an average of three  child  CWEs and a maximum of five. A few examples are provided in Figure~\ref{fig:predictionhierarchy}. When reporting performance, we take different top $k_i$ values of CWEs  from each level. The choice ($k_1=1, k_2=1, k_3=1$) gives precise prediction with only one path in the hierarchy. With moderate precision ($k_1=3, k_2=2, k_3=1$), there are  at most six possible paths. Finally, a more relaxed prediction can be obtained with ($k_1=5, k_2=2, k_3=2$), with at most twenty paths. If the true CWE(s) are present along the predicted paths, the prediction  is considered to be accurate.
Additionally we use the $F_1$-score of correctly classified links to evaluate the link prediction performance.
Table~\ref{tab:notation} lists the key notations used in the section.
\begin{table}[htpb]
\caption{Key notations used in the section}
\centering
\resizebox{0.9\linewidth}{!}{
\begin{tabular}{l|l}
\hline
\textbf{Notation}        & \textbf{Meaning}                                     \\ \hline
BERT\textsubscript{BASE} & Original pre-trained BERT model~\citep{devlin2018bert}                      \\
BERT\textsubscript{CVE} & Additional pre-training with CVE/CWE descriptions                               \\
LP                       & Link Prediction component only \\
LP+RD                   & Link Prediction coupled with Reconstruction Decoder        \\
\textsc{V2W-BERT}  & LP+RD, with BERT\textsubscript{CVE}                                        \\ \hline
$>n$                     & CVEs from CWEs with more than $n$ training instances \\
$[n_1, n_2]$            & CVEs from CWEs with training instances between $n_1$ to $n_2$        \\
$(k_1, k_2, k_3)$       & Top $k_1, k_2, k_3$ predictions for the $k_i$-th level in the hierarchy\\
Test 1      & Test instances from 2018 (near-future)\\
Test 2      & Test instances from 2019-2020 (far-future) \\
Link                     & Formulated as link prediction problem                \\
Class                    & Formulated as classification problem                 \\ \hline
\end{tabular}
}
\label{tab:notation}
\vspace{-0.4cm}
\end{table}
\subsection{Ablation Study}
\label{sec:ablation}
We evaluate each component of the \textsc{V2W-BERT} framework to find the best configuration for solving the problem. Additionally, we 
show how preserving the pre-trained BERT context using Reconstruction Decoder (RD) improves classification performance in rare and unseen cases.
The temporal partition of the dataset is used for evaluation.
\newline\newline
\noindent\textbf{Pooling and Combine Operations}\\
Experimental results show that \texttt{MEAN}-Pooling works best among the \texttt{CLS}, \texttt{MEAN}, and \texttt{MAX} pooling operations. When combining the vector representations of a CVE and CWE, concatenation of the absolute difference and multiplication  $(|\vx_{cve}- \vy_{cwe}|,\vx_{cve}\times \vy_{cwe})$ performs best, and these two operations are used for further experimentation. Due to page limitations, comparative details of different combination and pooling operations are given in Appendix~\ref{subsec:combination} and ~\ref{subsec:pooling} respectively.
\newline\newline
\noindent\textbf{Unsupervised Pre-training and Reconstruction Decoder}\\
To highlight the contribution of each component, we train \textsc{V2W-BERT} using only Link Prediction (LP) module with BERT\textsubscript{BASE} as a pre-trained model. This establishes our baseline for comparing the performance of additional pre-training and Reconstruction Decoder (RD).
Next, we fine-tune BERT\textsubscript{BASE} with all labeled and unlabeled CVE/CWE descriptions in the training years and train LP using this updated model. We refer this updated BERT model as BERT\textsubscript{CVE}.
Finally, we have a third experiment that uses LP and RD together using BERT\textsubscript{CVE} as a pre-trained model.

\begin{figure}[htpb]
    \includegraphics[width=\linewidth]{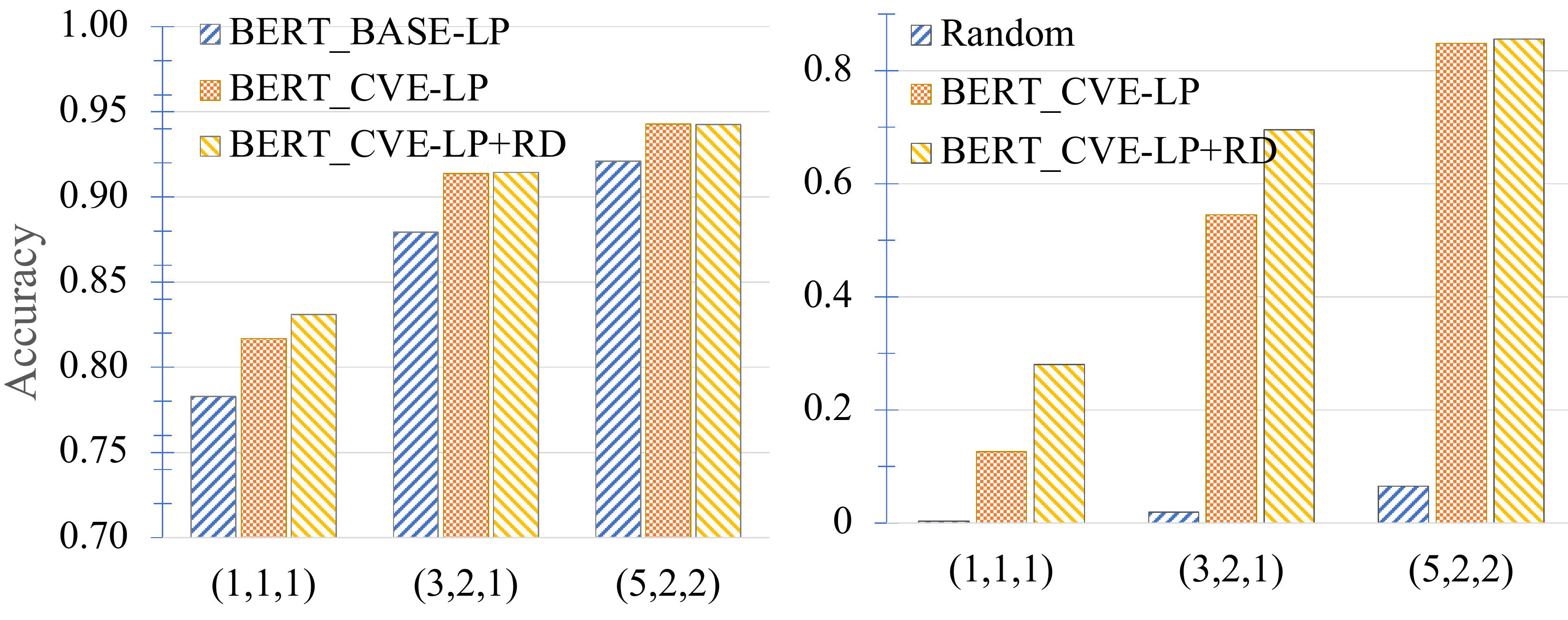}
    \caption{
     Precise and relaxed prediction accuracy for different components of \textsc{V2W-BERT}. {\em Left:} All data. {\em Right:} CVEs with unseen (zero-shot) CWEs.}
    \label{fig:ablationperformance}
\end{figure}

Fig~\ref{fig:ablationperformance} shows precise and relaxed prediction accuracy of cases mentioned above.
The use of BERT\textsubscript{CVE} outperforms BERT\textsubscript{BASE} in both the near and far future as learned cyber-security contexts help to transfer domain knowledge better.
The addition of the Reconstruction Decoder (RD) component helps preserve the context of BERT\textsubscript{CVE}, which improves performance in classifying CVEs of rare and unknown CWE classes, thus improving overall performance.
Test 2 has a lower accuracy than Test 1 as we predict two years into the future, containing different descriptions' style.
Appendix~\ref{subsec:pretrainrd} shows the quantitative details of these experiments.
\vspace{-0.4cm}
\newline\newline\noindent\textbf{Reconstruction Decoder for Few/Zero-shot Learning}\\
The Reconstruction Decoder (RD) component helps preserve the context of BERT\textsubscript{CVE}, which improves performance in classifying CVEs of rare and unknown CWE classes. 
We evaluate LP with and without the RD to highlight the improvement. 
We consider the CVEs of CWEs that appear in the test set but not in the training set or have few instances. We call these two cases zero-shot and few-shot, respectively. We use BERT\textsubscript{CVE} as the pre-trained model for experimentation.


\textit{\textbf{Zero-shot Performance:}}
We removed all CVEs of the descendants and ancestors of these unseen CWEs from the training process to avoid any bias for zero-shot evaluation.
Table~\ref{tab:zeroshotperformance} shows that the addition of Reconstruction Decoder (RD) improves the accuracy for unseen cases. The precise and relaxed prediction accuracies are evaluated for the CWEs that were absent during training. 
Here, ``Test 1 $(k_1,k_2,k_3)$, 89'' refers to 89 CVEs instances in year 2018 whose corresponding CWEs were unavailable during training.
The precise accuracy is relatively low but significantly higher than random prediction. For relaxed prediction, we get about $(86\%$ accuracy for Test 1 and $(61\%$  for Test 2 (illustrated in Figure~\ref{fig:ablationperformance}).
The performance of predicting unseen CVEs completely depends on inherent textual similarities between a CVE and CWE description. 

\begin{table}[htpb]
\caption{Zero-shot accuracy with and without RD}
\centering
\resizebox{\linewidth}{!}{
\begin{tabular}{l|ccc|ccc}
\hline
\multirow{2}{*}{\textbf{Model}} & \multicolumn{3}{c|}{\textbf{Test 1 $(k_1, k_2, k_3)$, 89}} & \multicolumn{3}{c}{\textbf{Test 2 $(k_1, k_2, k_3)$, 247}} \\ \cline{2-7} 
                  & \textbf{(1,1,1)} & \textbf{(3,2,1)} & \textbf{(5,2,2)} & \textbf{(1,1,1)} & \textbf{(3,2,1)} & \textbf{(5,2,2)} \\ \hline
Random & 0.0032           & 0.0196           & 0.0653           & 0.0032           & 0.0196           & 0.0653           \\
LP                & 0.1263           & 0.5454           & 0.8483           & 0.0273           & 0.2568           & 0.5902           \\
LP+RD             & \textbf{0.2809}  & \textbf{0.6954}  & \textbf{0.8558}  & \textbf{0.1012}  & \textbf{0.3475}  & \textbf{0.6104}  \\ \hline
\end{tabular}
}
\label{tab:zeroshotperformance}
\end{table}
\vspace{-0.3cm}
\textit{\textbf{Few-shot Performance:}}
Table~\ref{tab:fewshotperformance} shows the performance of CVEs where the corresponding CWEs have total training instances between ($[n_1, n_2]$). 
The ``Test 1, $n=[1,50]$, 1057'' refers to 1057 test CVE instances from 2018 whose corresponding CWEs had training examples between 1 to 50. 
With addition of RD, the model achieves significantly higher precise-prediction accuracy than Link Prediction (LP) alone.
The model achieves $71\%$-$84\%$ prediction accuracy in 2018 when we have only $51-100$ training instances in the past (1999-2017).
This improvement in rare cases is significant compared to related work, as detailed in \S\ref{sec:comparison}.


\begin{table}[htpb]
\caption{Few-shot accuracy evaluated for rare CWE classes with different training instances between $[n_1, n_2]$}
\centering
\resizebox{0.9\linewidth}{!}{
\begin{tabular}{c|ccc|ccc}
\hline
\textbf{Model}                   & \multicolumn{3}{c|}{\textbf{\begin{tabular}[c]{@{}c@{}}Test 1, n={[}1, 50{]}, 1057\end{tabular}}}   & \multicolumn{3}{c}{\textbf{\begin{tabular}[c]{@{}c@{}}Test 2, n={[}1, 50{]}, 2632\end{tabular}}}  \\ \hline
\textbf{$(k_1, k_2, k_3)$}  & \textbf{(1,1,1)}                      & \textbf{(3,2,1)}                    & \textbf{(5,2,2)}                   & \textbf{(1,1,1)}                      & \textbf{(3,2,1)}                    & \textbf{(5,2,2)}                    \\ \hline
LP & 0.2142                                & 0.4991                              & 0.671                              & 0.2462                                & 0.5151                              & 0.6306                              \\
LP+RD& \textbf{0.3199}                       & \textbf{0.6176}                     & \textbf{0.705}                     & \textbf{0.2474}                       & \textbf{0.5569}                     & \textbf{0.6736}                     \\ \hline
\multicolumn{1}{l|}{}      & \multicolumn{3}{c|}{\textbf{\begin{tabular}[c]{@{}c@{}}Test 1, n={[}51, 100{]}, 800\end{tabular}}}  & \multicolumn{3}{c}{\textbf{\begin{tabular}[c]{@{}c@{}}Test 2, n={[}51, 100{]}, 1221\end{tabular}}}  \\ \hline
LP   & 0.5687                                & 0.8075                              & \textbf{0.8400}                    & 0.5652                                & 0.7771                              & \textbf{0.8054}                     \\
LP+RD & \textbf{0.7087}                       & \textbf{0.8087}                     & 0.8375                             & \textbf{0.6457}                       & \textbf{0.7870}                     & 0.8035                              \\ \hline
\textbf{}                   & \multicolumn{3}{c|}{\textbf{\begin{tabular}[c]{@{}c@{}}Test 1, n={[}101, 150{]}, 690\end{tabular}}} & \multicolumn{3}{c}{\textbf{\begin{tabular}[c]{@{}c@{}}Test 2, n={[}101, 150{]}, 1643\end{tabular}}} \\ \hline
LP   & 0.6645                                & 0.8373                              & 0.9097                             & 0.4221                                & 0.6605                              & 0.7639                              \\
LP+RD & \textbf{0.7238}                       & \textbf{0.8475}                     & \textbf{0.9222}                    & \textbf{0.5091}                       & \textbf{0.6648}                     & \textbf{0.7849}                     \\ \hline
\end{tabular}
}
\vspace{-0.5cm}
\label{tab:fewshotperformance}
\end{table}
\subsection{Comparison with Related Approaches}
\label{sec:comparison}
\begin{figure}[htpb]
    \includegraphics[width=\linewidth]{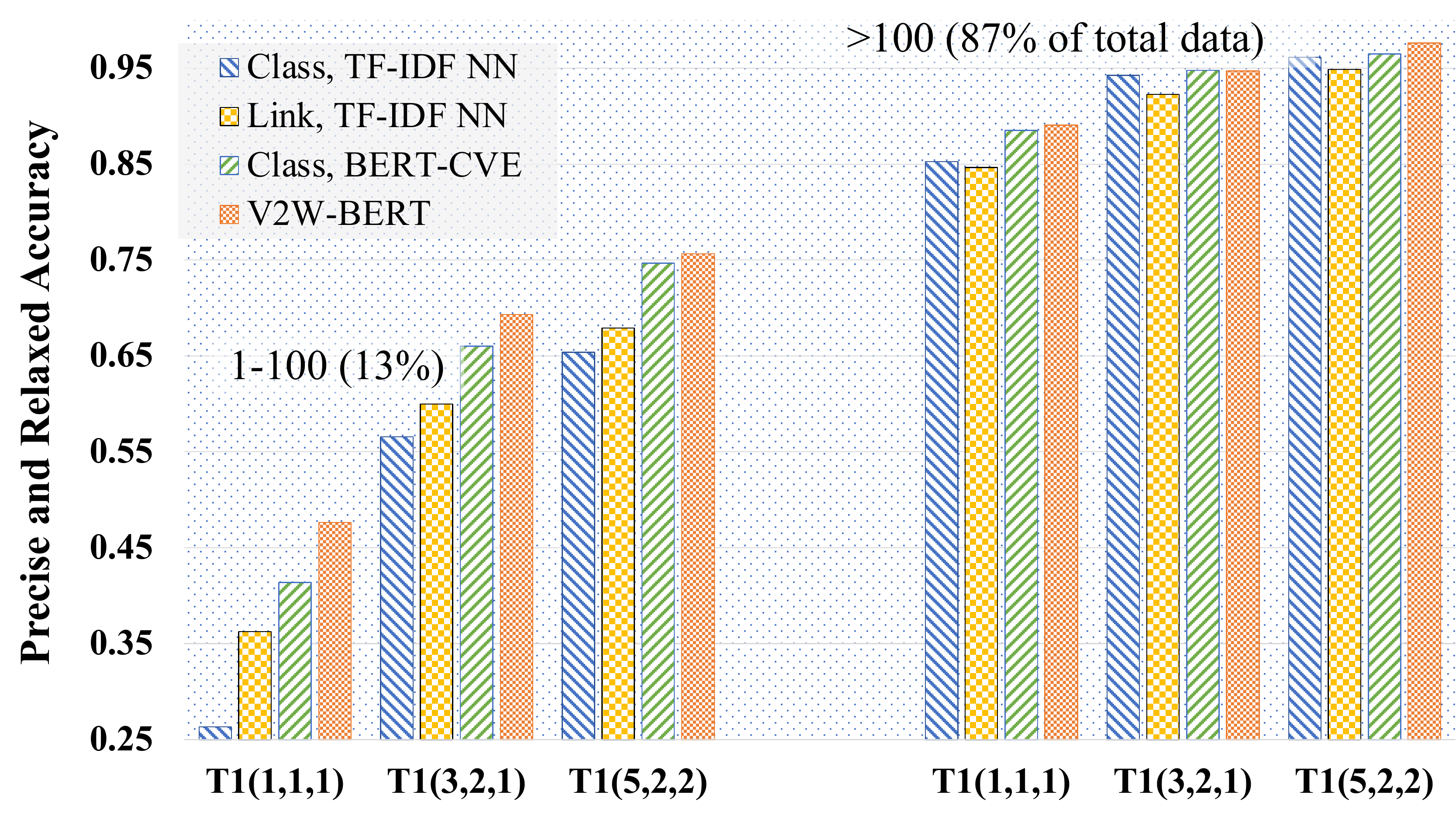}
    \caption{A summary of the key results for Test 1 (T1) showing superior performance of \textsc{V2W-BERT} with respect to other approaches, especially for rare CWEs classes. Details are provided in Table~\ref{tab:spatialpartition}.}
    \label{fig:temporalPerformance}
\end{figure}

We compare the performance of the \textsc{V2W-BERT} framework (using settings from \S\ref{sec:ablation}) with related work.
\textsc{V2W-BERT} is compared against two classification methods and a link association approach similar to ours. 
We compare with two classification approaches,  a TF-IDF based Neural Network (NN)~\citep{aghaei2019threatzoom} and a fine-tuned BERT classifier (this work).  
While fine-tuning the BERT classifier, we use the same pre-trained BERT\textsubscript{CVE} algorithm and  $\mathit{MEAN}$-Pooling as with \textsc{V2W-BERT}. Custom layers with dropout and fully connected Neural Networks are added on top of the pooling layer to predict all usable CWEs.
Additionally, we implement a TF-IDF feature-based link association method to train the model $F_\theta$. We use the TF-IDF feature directly and use the same $(|\vx_{cve}- \vy_{cwe}|,\vx_{cve}\times \vy_{cwe})$ combination operation and classification layer as we did in \textsc{V2W-BERT}. The training links are also kept same as \textsc{V2W-BERT}. 
We highlight the classification and link prediction based method with prefix `Class' and `Link' in the table.
\newline\newline
\noindent\textbf{Performance in the random partition of the dataset}\\
Table~\ref{tab:randomsplit} shows the comparative performance of the related methods. 
We take $70\%$ of the data for training from each category, $10\%$ for validation for hyper-parameter settings,  and $20\%$ for testing.
With more training data and examples overlapping all years, \textsc{V2W-BERT} 
and achieves  $\mathbf{89\%-97\%}$ precise and relaxed prediction accuracies.
\begin{table}[htpb]
\caption{Performance with randomly partitioned dataset}
\centering
\resizebox{0.7\linewidth}{!}{
\begin{tabular}{l|lll}
\hline
\multicolumn{1}{c|}{\multirow{2}{*}{\textbf{Model}}} &
  \multicolumn{3}{c}{\textbf{Test Set $(k_1, k_2, k_3)$}} \\ \cline{2-4} 
\multicolumn{1}{c|}{} &
  \multicolumn{1}{c}{\textbf{(1,1,1)}} &
  \multicolumn{1}{c}{\textbf{(3,2,1)}} &
  \multicolumn{1}{c}{\textbf{(5,2,2)}} \\ \hline
Class, TF-IDF NN            & 0.8606          & 0.9464          & 0.9668          \\
Link, TF-IDF NN             & 0.8642          & 0.9502          & 0.9693          \\
Class, BERT\textsubscript{CVE} & 0.8812          & 0.9503          & 0.9689          \\
Link, \textsc{V2W-BERT}                   & \textbf{0.8916} & \textbf{0.9523} & \textbf{0.9723} \\ \hline
\end{tabular}
}
\label{tab:randomsplit}
\vspace{-0.4cm}
\end{table}
\newline\newline \noindent\textbf{Performance in the temporal partition of the dataset}\\
Unlike random partition, where we have taken training examples from each category, temporal partition is more challenging and reflective of the application.
Table~\ref{tab:spatialpartition} compares the accuracy of \textsc{V2W-BERT} trained with data from 1999-2017, and tested for 2018 (Test 1) and 2019-2020 (Test 2). Key results are illustrated in Figure~\ref{fig:temporalPerformance}. 
To highlight the performance of CVEs of rare and frequently occurring CWEs, we split the test sets by CWEs having $1-100$ training examples,  and by CWEs  with more than a hundred training examples.
The \textsc{V2W-BERT} outperforms the competing approaches in both precise and relaxed predictions,  overall as well as in rare and frequently occurring cases.
For CWEs with $\geq 100$ training instances, \textsc{V2W-BERT} achieves $\mathbf{89\%-98\%}$ precise and relaxed prediction accuracy in Test 1 (2018).
The performance on Test 2 data is lower than that of Test 1, since the former is further into the future. To demonstrate sustainability of \textsc{V2W-BERT}, we experimented by adding recent data (from 2018) for training, and it improves the performance on Test 2 data (Appendix~\ref{subsec:additionaltraining}).
\begin{table}[htpb]
\caption{Performance comparison of \textsc{V2W-BERT}}
\centering
\resizebox{\linewidth}{!}{
\begin{tabular}{l|l|lll|lll}
\hline
\multirow{2}{*}{\textbf{\begin{tabular}[c]{@{}l@{}}\end{tabular}}} & \multicolumn{1}{c|}{\multirow{2}{*}{\textbf{Model}}} & \multicolumn{3}{c|}{\textbf{Test 1 $(k_1, k_2, k_3)$}} & \multicolumn{3}{c}{\textbf{Test 2 $(k_1, k_2, k_3)$}} \\ \cline{3-8} 
 & \multicolumn{1}{c|}{} & \multicolumn{1}{c}{\textbf{(1,1,1)}} & \multicolumn{1}{c}{\textbf{(3,2,1)}} & \multicolumn{1}{c|}{\textbf{(5,2,2)}} & \multicolumn{1}{c}{\textbf{(1,1,1)}} & \multicolumn{1}{c}{\textbf{(3,2,1)}} & \multicolumn{1}{c}{\textbf{(5,2,2)}} \\ \hline

\multirow{4}{*}{1-100} & Class, TF-IDF NN & 0.2631 & 0.5656 & 0.6537 & 0.2519 & 0.4838 & 0.5739 \\
 & Link, TF-IDF NN & 0.3626 & 0.5998 & 0.6791 & 0.3395 & 0.564 & 0.659 \\
 & Class, BERT\textsubscript{CVE} & 0.4138 & 0.6602 & 0.7466 & 0.2914 & 0.6105 & 0.6902 \\
 & Link, \textsc{V2W-BERT} & \textbf{0.4765} & \textbf{0.6933} & \textbf{0.7564} & \textbf{0.4072} & \textbf{0.6293} & \textbf{0.7179} \\ \hline

\multirow{4}{*}{\textgreater{}100} & Class, TF-IDF NN & 0.8524 & 0.9425 & 0.9616 & 0.7815 & 0.8953 & 0.9404 \\
 & Link, TF-IDF NN & 0.8463 & 0.9227 & 0.9485 & 0.7604 & 0.8738 & 0.9153 \\
 & Class, BERT\textsubscript{CVE} & 0.8852 & 0.9479 & 0.9649 & 0.8067 & 0.9064 & 0.9414 \\
 & Link, \textsc{V2W-BERT} & \textbf{0.8905} & \textbf{0.947} & \textbf{0.9763} & \textbf{0.8113} & \textbf{0.9123} & \textbf{0.9492} \\ \hline

\multirow{4}{*}{All} & Class, TF-IDF NN            & 0.775  & 0.893  & 0.9298 & 0.6886 & 0.8231 & 0.8761 \\
& Link, TF-IDF NN             & 0.7828 & 0.8803 & 0.9132 & 0.6863 & 0.8196 & 0.8706 \\
& Class, BERT\textsubscript{CVE} & 0.8232 & 0.9101 & 0.9363 & 0.7163 & 0.8578 & 0.9038 \\
& Link, \textsc{V2W-BERT}                                             & \textbf{0.8362}  & \textbf{0.914}   &  \textbf{0.9442}  & \textbf{0.7345}  & \textbf{0.8594}  & \textbf{0.9151}  \\\hline

\end{tabular}
}
\label{tab:spatialpartition}
\end{table}

\textit{\textbf{ $\mathbf{F_1}$-Score of predicted links:}} We evaluate both link and unlink pairs that are correctly classified. Only the two link-based methods (\textsc{V2W-BERT} and Link, TF-IDF NN) predict links.
\textsc{V2W-BERT} achieves $F_1$-Scores of $0.93$  for Test 1,  and $0.92$ for Test 2, where as TF-IDF NN achieves $0.91$ and $0.88$ respectively (\S\ref{subsec:linkpredperformance}).
Performance of predicting links is higher than the precise CWE predictions since predicting a CWE accurately down to the leaf node requires all links to the ancestor to be correctly predicted.
\newline\indent
\textit{\textbf{Zero-shot performance of link methods:}}  Table~\ref{tab:linkzeroshotperformance} captures classification performance of CVEs associated with CWEs not seen  in training. Only the link-based methods are compared since classification-based approaches do not support this task. The link-based TF-IDF NN performs worse than random choice since it is over-fitted to the available training CWEs.
\begin{table}[htpb]
\caption{Zero-shot accuracy of link-based methods}
\centering
\resizebox{1\linewidth}{!}{
\begin{tabular}{l|ccc|ccc}
\hline
\multirow{2}{*}{\textbf{Model}} & \multicolumn{3}{c|}{\textbf{Test 1 $(k_1, k_2, k_3)$, 89}} & \multicolumn{3}{c}{\textbf{Test 2 $(k_1, k_2, k_3)$, 247}}  \\ \cline{2-7} 
                                & \textbf{(1,1,1)} & \textbf{(3,2,1)} & \textbf{(5,2,2)} & \textbf{(1,1,1)} & \textbf{(3,2,1)} & \textbf{(5,2,2)} \\ \hline
Random& 0.0032 & 0.0196 & 0.0653 & 0.0032 & 0.0196 & 0.0653 \\
Link, TF-IDF NN   & 0.0000 & 0.1158 & 0.4875 & 0.0000 & 0.0562 & 0.1717 \\
Link, \textsc{V2W-BERT}   & \textbf{0.2809}  & \textbf{0.6954}  & \textbf{0.8558}  & \textbf{0.1012}  & \textbf{0.3475}  & \textbf{0.6104}  \\ \hline
\end{tabular}
}
\label{tab:linkzeroshotperformance}
\vspace{-0.5cm}
\end{table}
\newline\newline
\noindent\textbf{Predicting a new CWE definition}\\
For a given CVE, \textsc{V2W-BERT} gives link and unlink values to all available CWEs. 
If the link value is higher than unlink, we consider the CVE to be associated with that CWE. The link value represents the confidence about the association of a vulnerability to a weakness. We can push this confidence boundary for a more robust prediction and consider the link only if the value is greater than a threshold $\beta$.
For a CVE description, if all link values to the available CWEs are less than $\beta$, then the CVE description has a different style, or we need a new CWE definition. 
Appendix~\ref{subsec:cweprediction} shows experimental evidence where we get most occurrences of all unlinks in the case of unseen CWEs.

\section{Summary and Future Work}
\label{sec:conclusions}
We presented a Transformer-based framework (\textsc{V2W-BERT}) to efficiently map CVEs (specific vulnerability reports) to hierarchically structured CWEs (weakness descriptions). Using data from standard sources, we demonstrated high quality results that outperform previous efforts. We also demonstrated that our approach not only performs well for CWE classes with abundant data, but also for rare CWE classes with little or no data to train.
Since classifying rare CWEs has been an explored problem in literature, our framework provides a promising novel approach towards a viable practical solution to efficiently classify increasing more and diverse software vulnerabilities. 
We also demonstrated that our framework can learn from historic data and predict new information that has not been seen before.
Our future work will focus on scaling larger pre-trained BERT models with high-performance computing platforms to further enhance the classification performance, and automated suggestions for defining new weaknesses to match novel vulnerabilities.

\bibliographystyle{ACM-Reference-Format}
\bibliography{halappanavar, sidcve, sidgnn}


\begin{thebibliography}{21}


\ifx \showCODEN    \undefined \def \showCODEN     #1{\unskip}     \fi
\ifx \showDOI      \undefined \def \showDOI       #1{#1}\fi
\ifx \showISBNx    \undefined \def \showISBNx     #1{\unskip}     \fi
\ifx \showISBNxiii \undefined \def \showISBNxiii  #1{\unskip}     \fi
\ifx \showISSN     \undefined \def \showISSN      #1{\unskip}     \fi
\ifx \showLCCN     \undefined \def \showLCCN      #1{\unskip}     \fi
\ifx \shownote     \undefined \def \shownote      #1{#1}          \fi
\ifx \showarticletitle \undefined \def \showarticletitle #1{#1}   \fi
\ifx \showURL      \undefined \def \showURL       {\relax}        \fi
\providecommand\bibfield[2]{#2}
\providecommand\bibinfo[2]{#2}
\providecommand\natexlab[1]{#1}
\providecommand\showeprint[2][]{arXiv:#2}

\bibitem[\protect\citeauthoryear{Aghaei and Al-Shaer}{Aghaei and
  Al-Shaer}{2019}]%
        {aghaei2019threatzoom}
\bibfield{author}{\bibinfo{person}{Ehsan Aghaei} {and} \bibinfo{person}{Ehab
  Al-Shaer}.} \bibinfo{year}{2019}\natexlab{}.
\newblock \showarticletitle{ThreatZoom: neural network for automated
  vulnerability mitigation}. In \bibinfo{booktitle}{\emph{Proceedings of the
  6th Annual Symposium on Hot Topics in the Science of Security}}.
  \bibinfo{pages}{1--3}.
\newblock


\bibitem[\protect\citeauthoryear{Agirre, Banea, et~al\mbox{.}}{Agirre
  et~al\mbox{.}}{2015}]%
        {agirre2015semeval}
\bibfield{author}{\bibinfo{person}{Eneko Agirre}, \bibinfo{person}{Carmen
  Banea}, {et~al\mbox{.}}} \bibinfo{year}{2015}\natexlab{}.
\newblock \showarticletitle{Semeval-2015 task 2: Semantic textual similarity,
  english, spanish and pilot on interpretability}. In
  \bibinfo{booktitle}{\emph{Proceedings of the 9th International Workshop on
  Semantic Evaluation (SemEval 2015)}}. \bibinfo{pages}{252--263}.
\newblock


\bibitem[\protect\citeauthoryear{Aota, Kanehara, Kubo, Murata, Sun, and
  Takahashi}{Aota et~al\mbox{.}}{2020}]%
        {aota2020automation}
\bibfield{author}{\bibinfo{person}{Masaki Aota}, \bibinfo{person}{Hideaki
  Kanehara}, \bibinfo{person}{Masaki Kubo}, \bibinfo{person}{Noboru Murata},
  \bibinfo{person}{Bo Sun}, {and} \bibinfo{person}{Takeshi Takahashi}.}
  \bibinfo{year}{2020}\natexlab{}.
\newblock \showarticletitle{Automation of Vulnerability Classification from its
  Description using Machine Learning}. In \bibinfo{booktitle}{\emph{2020 IEEE
  Symposium on Computers and Communications (ISCC)}}. IEEE,
  \bibinfo{pages}{1--7}.
\newblock


\bibitem[\protect\citeauthoryear{Cer, Yang, Kong, Hua, Limtiaco, John,
  Constant, Guajardo-Cespedes, Yuan, Tar, et~al\mbox{.}}{Cer
  et~al\mbox{.}}{2018}]%
        {cer2018universal}
\bibfield{author}{\bibinfo{person}{Daniel Cer}, \bibinfo{person}{Yinfei Yang},
  \bibinfo{person}{Sheng-yi Kong}, \bibinfo{person}{Nan Hua},
  \bibinfo{person}{Nicole Limtiaco}, \bibinfo{person}{Rhomni~St John},
  \bibinfo{person}{Noah Constant}, \bibinfo{person}{Mario Guajardo-Cespedes},
  \bibinfo{person}{Steve Yuan}, \bibinfo{person}{Chris Tar}, {et~al\mbox{.}}}
  \bibinfo{year}{2018}\natexlab{}.
\newblock \showarticletitle{Universal Sentence Encoder}.
\newblock \bibinfo{journal}{\emph{arXiv:1803.11175}} (\bibinfo{year}{2018}).
\newblock


\bibitem[\protect\citeauthoryear{Chicco}{Chicco}{2020}]%
        {chicco2020siamese}
\bibfield{author}{\bibinfo{person}{Davide Chicco}.}
  \bibinfo{year}{2020}\natexlab{}.
\newblock \showarticletitle{Siamese neural networks: An overview}.
\newblock \bibinfo{journal}{\emph{Artificial Neural Networks}}
  (\bibinfo{year}{2020}), \bibinfo{pages}{73--94}.
\newblock


\bibitem[\protect\citeauthoryear{Cruzes, Felderer, Oyetoyan, Gander, and
  Pekaric}{Cruzes et~al\mbox{.}}{2017}]%
        {Cruzes2017}
\bibfield{author}{\bibinfo{person}{Daniela~Soares Cruzes},
  \bibinfo{person}{Michael Felderer}, \bibinfo{person}{Tosin~Daniel Oyetoyan},
  \bibinfo{person}{Matthias Gander}, {and} \bibinfo{person}{Irdin Pekaric}.}
  \bibinfo{year}{2017}\natexlab{}.
\newblock \showarticletitle{How is Security Testing Done in Agile Teams? {A}
  Cross-Case Analysis of Four Software Teams}. In
  \bibinfo{booktitle}{\emph{{XP}}} \emph{(\bibinfo{series}{Lecture Notes in
  Business Information Processing}, Vol.~\bibinfo{volume}{283})}.
  \bibinfo{pages}{201--216}.
\newblock


\bibitem[\protect\citeauthoryear{Devlin, Chang, Lee, and Toutanova}{Devlin
  et~al\mbox{.}}{2018}]%
        {devlin2018bert}
\bibfield{author}{\bibinfo{person}{Jacob Devlin}, \bibinfo{person}{Ming-Wei
  Chang}, \bibinfo{person}{Kenton Lee}, {and} \bibinfo{person}{Kristina
  Toutanova}.} \bibinfo{year}{2018}\natexlab{}.
\newblock \showarticletitle{Bert: Pre-training of deep bidirectional
  transformers for language understanding}.
\newblock \bibinfo{journal}{\emph{arXiv:1810.04805}} (\bibinfo{year}{2018}).
\newblock


\bibitem[\protect\citeauthoryear{Han, Li, Xing, Liu, and Feng}{Han
  et~al\mbox{.}}{2017}]%
        {han2017learning}
\bibfield{author}{\bibinfo{person}{Zhuobing Han}, \bibinfo{person}{Xiaohong
  Li}, \bibinfo{person}{Zhenchang Xing}, \bibinfo{person}{Hongtao Liu}, {and}
  \bibinfo{person}{Zhiyong Feng}.} \bibinfo{year}{2017}\natexlab{}.
\newblock \showarticletitle{Learning to predict severity of software
  vulnerability using only vulnerability description}. In
  \bibinfo{booktitle}{\emph{2017 IEEE International Conference on Software
  Maintenance and Evolution (ICSME)}}. IEEE, \bibinfo{pages}{125--136}.
\newblock


\bibitem[\protect\citeauthoryear{{Jimenez}, {Papadakis}, and {Traon}}{{Jimenez}
  et~al\mbox{.}}{2016}]%
        {Jimenez2016}
\bibfield{author}{\bibinfo{person}{M. {Jimenez}}, \bibinfo{person}{M.
  {Papadakis}}, {and} \bibinfo{person}{Y.~L. {Traon}}.}
  \bibinfo{year}{2016}\natexlab{}.
\newblock \showarticletitle{An Empirical Analysis of Vulnerabilities in OpenSSL
  and the Linux Kernel}. In \bibinfo{booktitle}{\emph{2016 23rd Asia-Pacific
  Software Engineering Conference (APSEC)}}. \bibinfo{pages}{105--112}.
\newblock
\showISSN{1530-1362}


\bibitem[\protect\citeauthoryear{Loshchilov and Hutter}{Loshchilov and
  Hutter}{2017}]%
        {loshchilov2017decoupled}
\bibfield{author}{\bibinfo{person}{Ilya Loshchilov} {and}
  \bibinfo{person}{Frank Hutter}.} \bibinfo{year}{2017}\natexlab{}.
\newblock \showarticletitle{Decoupled weight decay regularization}.
\newblock \bibinfo{journal}{\emph{arXiv:1711.05101}} (\bibinfo{year}{2017}).
\newblock


\bibitem[\protect\citeauthoryear{Lu, Jiao, and Zhang}{Lu et~al\mbox{.}}{2020}]%
        {lu2020twinbert}
\bibfield{author}{\bibinfo{person}{Wenhao Lu}, \bibinfo{person}{Jian Jiao},
  {and} \bibinfo{person}{Ruofei Zhang}.} \bibinfo{year}{2020}\natexlab{}.
\newblock \showarticletitle{TwinBERT: Distilling knowledge to twin-structured
  BERT models for efficient retrieval}.
\newblock \bibinfo{journal}{\emph{arXiv:2002.06275}} (\bibinfo{year}{2020}).
\newblock


\bibitem[\protect\citeauthoryear{Martin and Barnum}{Martin and Barnum}{2008}]%
        {Martin2008}
\bibfield{author}{\bibinfo{person}{Robert~A. Martin} {and}
  \bibinfo{person}{Sean Barnum}.} \bibinfo{year}{2008}\natexlab{}.
\newblock \showarticletitle{Common Weakness Enumeration {(CWE)} Status Update}.
\newblock \bibinfo{journal}{\emph{Ada Lett.}} (\bibinfo{year}{2008}),
  \bibinfo{pages}{88--91}.
\newblock


\bibitem[\protect\citeauthoryear{Na, Kim, and Kim}{Na et~al\mbox{.}}{2016}]%
        {na2016study}
\bibfield{author}{\bibinfo{person}{Sarang Na}, \bibinfo{person}{Taeeun Kim},
  {and} \bibinfo{person}{Hwankuk Kim}.} \bibinfo{year}{2016}\natexlab{}.
\newblock \showarticletitle{A study on the classification of common
  vulnerabilities and exposures using na{\"\i}ve bayes}. In
  \bibinfo{booktitle}{\emph{International Conference on Broadband and Wireless
  Computing, Communication and Applications}}. Springer,
  \bibinfo{pages}{657--662}.
\newblock


\bibitem[\protect\citeauthoryear{Nakagawa, Nagai, Kanehara, Furumoto, Takita,
  Shiraishi, Takahashi, Mohri, Takano, and Morii}{Nakagawa
  et~al\mbox{.}}{2019}]%
        {nakagawa2019character}
\bibfield{author}{\bibinfo{person}{Shunta Nakagawa}, \bibinfo{person}{Tatsuya
  Nagai}, \bibinfo{person}{Hideaki Kanehara}, \bibinfo{person}{Keisuke
  Furumoto}, \bibinfo{person}{Makoto Takita}, \bibinfo{person}{Yoshiaki
  Shiraishi}, \bibinfo{person}{Takeshi Takahashi}, \bibinfo{person}{Masami
  Mohri}, \bibinfo{person}{Yasuhiro Takano}, {and} \bibinfo{person}{Masakatu
  Morii}.} \bibinfo{year}{2019}\natexlab{}.
\newblock \showarticletitle{Character-level convolutional neural network for
  predicting severity of software vulnerability from vulnerability
  description}.
\newblock \bibinfo{journal}{\emph{IEICE Transactions on Information and
  Systems}} \bibinfo{volume}{102}, \bibinfo{number}{9} (\bibinfo{year}{2019}),
  \bibinfo{pages}{1679--1682}.
\newblock


\bibitem[\protect\citeauthoryear{Neuhaus and Zimmermann}{Neuhaus and
  Zimmermann}{2010}]%
        {neuhaus2010security}
\bibfield{author}{\bibinfo{person}{Stephan Neuhaus} {and}
  \bibinfo{person}{Thomas Zimmermann}.} \bibinfo{year}{2010}\natexlab{}.
\newblock \showarticletitle{Security trend analysis with cve topic models}. In
  \bibinfo{booktitle}{\emph{2010 IEEE 21st International Symposium on Software
  Reliability Engineering}}. IEEE, \bibinfo{pages}{111--120}.
\newblock


\bibitem[\protect\citeauthoryear{Rehman and Mustafa}{Rehman and
  Mustafa}{2012}]%
        {rehman2012software}
\bibfield{author}{\bibinfo{person}{Shabana Rehman} {and}
  \bibinfo{person}{Khurram Mustafa}.} \bibinfo{year}{2012}\natexlab{}.
\newblock \showarticletitle{Software design level vulnerability classification
  model}.
\newblock \bibinfo{journal}{\emph{International Journal of Computer Science and
  Security (IJCSS)}} \bibinfo{volume}{6}, \bibinfo{number}{4}
  (\bibinfo{year}{2012}), \bibinfo{pages}{238}.
\newblock


\bibitem[\protect\citeauthoryear{Reimers and Gurevych}{Reimers and
  Gurevych}{2019}]%
        {reimers2019sentence}
\bibfield{author}{\bibinfo{person}{Nils Reimers} {and} \bibinfo{person}{Iryna
  Gurevych}.} \bibinfo{year}{2019}\natexlab{}.
\newblock \showarticletitle{Sentence-bert: Sentence embeddings using siamese
  bert-networks}.
\newblock \bibinfo{journal}{\emph{arXiv:1908.10084}} (\bibinfo{year}{2019}).
\newblock


\bibitem[\protect\citeauthoryear{Sun, Qiu, Xu, and Huang}{Sun
  et~al\mbox{.}}{2019}]%
        {sun2019fine}
\bibfield{author}{\bibinfo{person}{Chi Sun}, \bibinfo{person}{Xipeng Qiu},
  \bibinfo{person}{Yige Xu}, {and} \bibinfo{person}{Xuanjing Huang}.}
  \bibinfo{year}{2019}\natexlab{}.
\newblock \showarticletitle{How to fine-tune bert for text classification?}. In
  \bibinfo{booktitle}{\emph{China National Conference on Chinese Computational
  Linguistics}}. Springer, \bibinfo{pages}{194--206}.
\newblock


\bibitem[\protect\citeauthoryear{Vaswani, Shazeer, Parmar, Uszkoreit, Jones,
  Gomez, Kaiser, and Polosukhin}{Vaswani et~al\mbox{.}}{2017}]%
        {vaswani2017attention}
\bibfield{author}{\bibinfo{person}{Ashish Vaswani}, \bibinfo{person}{Noam
  Shazeer}, \bibinfo{person}{Niki Parmar}, \bibinfo{person}{Jakob Uszkoreit},
  \bibinfo{person}{Llion Jones}, \bibinfo{person}{Aidan~N Gomez},
  \bibinfo{person}{{\L}ukasz Kaiser}, {and} \bibinfo{person}{Illia
  Polosukhin}.} \bibinfo{year}{2017}\natexlab{}.
\newblock \showarticletitle{Attention is all you need}. In
  \bibinfo{booktitle}{\emph{Advances in Neural Information Processing Systems
  (NIPS}}. \bibinfo{pages}{5998--6008}.
\newblock


\bibitem[\protect\citeauthoryear{Wagner, Mahbub, Palomar, and Abdallah}{Wagner
  et~al\mbox{.}}{2019}]%
        {WAGNER2019101589}
\bibfield{author}{\bibinfo{person}{Thomas~D. Wagner}, \bibinfo{person}{Khaled
  Mahbub}, \bibinfo{person}{Esther Palomar}, {and} \bibinfo{person}{Ali~E.
  Abdallah}.} \bibinfo{year}{2019}\natexlab{}.
\newblock \showarticletitle{Cyber threat intelligence sharing: Survey and
  research directions}.
\newblock \bibinfo{journal}{\emph{Computers \& Security}}  \bibinfo{volume}{87}
  (\bibinfo{year}{2019}), \bibinfo{pages}{101589}.
\newblock
\urldef\tempurl%
\url{https://doi.org/10.1016/j.cose.2019.101589}
\showDOI{\tempurl}


\bibitem[\protect\citeauthoryear{Wei and Zou}{Wei and Zou}{2019}]%
        {wei2019eda}
\bibfield{author}{\bibinfo{person}{Jason Wei} {and} \bibinfo{person}{Kai Zou}.}
  \bibinfo{year}{2019}\natexlab{}.
\newblock \showarticletitle{Eda: Easy data augmentation techniques for boosting
  performance on text classification tasks}.
\newblock \bibinfo{journal}{\emph{arXiv:1901.11196}} (\bibinfo{year}{2019}).
\newblock


\end{thebibliography}


\clearpage\newpage
\appendix
\section{Appendix}
In the appendix, we discuss in more detail some components of the V2W\textsubscript{BERT} framework. 








\subsection{Masked Language Model for Pre-training}
\label{subsec:pretrainarchitecture}
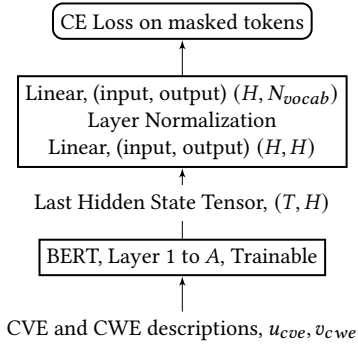
\begin{figure}[htbp]
    \centering
    \resizebox{0.6\linewidth}{!}{
    \tikzset{
    block/.style={
      draw, 
      rectangle, 
      thick,
      minimum height=0.5cm, 
      minimum width=3cm, align=center
      }, 
    line/.style={->,>=latex'},
    cloud/.style={
      draw=red,
      thick,
      ellipse,
      fill=red!20,
      minimum height=1em
    },
    roundblock/.style={
      draw, 
      rectangle, 
      rounded corners,
      thick,
      minimum height=0.5cm, 
      minimum width=3cm, align=center
      }, 
    }
    \begin{tikzpicture}[node distance=2.0cm]

\node (cvedescription) {CVE and CWE descriptions, $u_{cve}, v_{cwe}$};

\node[block, above =0.5cm of cvedescription] (bert) {BERT, Layer 1 to $A$, Trainable};

\node[block, above =1.0cm of bert] (lmlinear)
{
Linear, (input, output) $(H, N_{vocab})$\\
Layer Normalization\\
Linear, (input, output) $(H, H)$
};
\node[roundblock, above =0.5cm of lmlinear] (lmloss) {CE Loss on masked tokens};

\draw[line] (cvedescription.north) -- (bert.south);
\draw[line] (bert.north) -- (lmlinear.south) node [midway,fill=white]{Last Hidden State Tensor, $(T, H)$};
\draw[line] (lmlinear.north) -- (lmloss.south);

\end{tikzpicture}
}
\caption{Architecture of Masked Language Model.}
\label{fig:pretrainedmodel}
\end{figure}

Fig~\ref{fig:pretrainedmodel} shows a simplistic view of fine-tuning BERT with Masked LM. We allow all layers of BERT to update in this step as we are learning  the relevant cyber-security context. A custom Language Model (LM) layer is added on top of the BERT encoder, which takes the last hidden state tensor from the BERT encoder and then passes that to a linear layer of input-output size $(H, H)$. Then layer normalization is performed, and values are passed to a  linear layer with an input-output feature size $(H, N_{{\rm vocab}})$ to predict masked tokens. The cross-entropy loss is used on the predicted masked tokens to optimize the model.

\subsection{Link Prediction (LP) with Different Combination Operations}
\label{subsec:combination}
Following  recent work~\citep{cer2018universal,reimers2019sentence}, the V2W\textsubscript{BERT} is evaluated by different combination operations. For simplicity, only the Link Prediction (LP) component is used with \texttt{CLS}-pooling. The BERT\textsubscript{BASE}  is used as the pre-trained model for experimentation, and experiments are run for ten epochs only.

Table~\ref{tab:combination} shows comparative performance of some combination operations. The concatenation operation $(\vx, \vy)$ does not achieve good performance, but  multiplication, $(\vx\times \vy)$,  performs better than absolute difference, $(|\vx- \vy|)$. Their combination  $(|\vx- \vy|,\vx\times \vy)$  shows the overall best performance, and is used for further experiments.

\begin{table}[htbp]
\caption{Accuracy of Link Prediction (LP) component over different combination operations.}
\centering
\resizebox{1\linewidth}{!}{
\begin{tabular}{l|ccc|ccc}
\hline
\multicolumn{1}{c|}{\multirow{2}{*}{\textbf{Combination}}} & \multicolumn{3}{c|}{\textbf{Test 1 $(k_1, k_2, k_3)$}} & \multicolumn{3}{c}{\textbf{Test 2 $(k_1, k_2, k_3)$}} \\ \cline{2-7} 
\multicolumn{1}{c|}{}                                & \textbf{(1,1,1)} & \textbf{(3,2,1)} & \textbf{(5,2,2)} & \textbf{(1,1,1)} & \textbf{(3,2,1)} & \textbf{(5,2,2)} \\ \hline
$(\vx, \vy)$                                            & 0.2631           & 0.5401           & 0.6517           & 0.2237           & 0.5063           & 0.6288                                \\
$(|\vx -  \vy|)$                                        & 0.6471           & 0.8816           & 0.9175           & 0.544            & 0.8237           & 0.8742                                \\
$(\vx\times \vy)$                                       & 0.7657           & \textbf{0.8885}  & \textbf{0.9279}  & 0.6897           & 0.8395           & \textbf{0.8953}                       \\
$(|\vx -  \vy|,\vx\times \vy)$                  & \textbf{0.7829}  & 0.8794           & 0.9209           & \textbf{0.6995}  & 0.8337           & 0.8879                                \\
$(\vx, \vy,\vx\times   \vy)$                    & 0.7628           & 0.8846           & 0.9225           & 0.6915           & \textbf{0.8411}  & 0.8880                                \\
$(\vx, \vy,|\vx- \vy|)$                         & 0.7769           & 0.8828           & 0.9233           & 0.6839           & 0.822            & 0.8823                                \\
$(\vx, \vy,|\vx-   \vy|,\vx\times \vy)$ & 0.7815           & 0.8827           & 0.9211           & 0.6833           & 0.8203           & 0.8766                                \\ \hline
\end{tabular}
}
\label{tab:combination}
\end{table}

\subsection{Link Prediction (LP) with different Pooling operations}
\label{subsec:pooling}
Reimers et. al.~\citep{reimers2019sentence} have shown that other pooling operations can outperform \texttt{CLS}-Pooling. In this work, we have investigated V2W\textsubscript{BERT} with three pooling operations, \texttt{CLS}-pooling, \texttt{MAX}-pooling, and \texttt{MEAN}-pooling. Table~\ref{tab:pooling} shows comparative performance of different BERT poolers with $(|\vx_{cve}- \vy_{cwe}|,\vx_{cve}\times \vy_{cwe})$ as the combination operation. BERT\textsubscript{BASE} is used as the pre-trained model and the experiments are run for ten epochs only. 
\texttt{MEAN}-pooling has shown marginally better performance than \texttt{CLS}-Pooling, and is used for V2W\textsubscript{BERT}.

\begin{table}[htbp]
\caption{Accuracy of Link Prediction (LP) component over different pooling approaches.}
\centering
\resizebox{\linewidth}{!}{
\begin{tabular}{l|ccc|ccc}
\hline
\multicolumn{1}{c|}{\multirow{2}{*}{\textbf{Pooling}}} & \multicolumn{3}{c|}{\textbf{Test 1 $(k_1, k_2, k_3)$}} & \multicolumn{3}{c}{\textbf{Test 2 $(k_1, k_2, k_3)$}} \\ \cline{2-7} 
\multicolumn{1}{c|}{}                                & \textbf{(1,1,1)} & \textbf{(3,2,1)} & \textbf{(5,2,2)} & \textbf{(1,1,1)} & \textbf{(3,2,1)} & \textbf{(5,2,2)} \\ \hline
CLS-Pooling                       & \textbf{0.7829}  & 0.8794           & 0.9209           & \textbf{0.6995}  & 0.8337           & 0.8879                                \\
MAX-Pooling                       & 0.7592           & 0.872            & 0.9175           & 0.6705           & 0.818            & 0.8748                                \\
MEAN-Pooling                      & 0.782            & \textbf{0.8886}  & \textbf{0.9244}  & 0.6874           & \textbf{0.8364}  & \textbf{0.8897}                       \\\hline
\end{tabular}
}
\label{tab:pooling}
\end{table}

\subsection{Link Prediction (LP) and Reconstruction Decoder (RD) with different pre-trained models.}
\label{subsec:pretrainrd}
Table~\ref{tab:pretrained} shows precise and relaxed prediction accuracy of the three scenarios of \textsc{V2W-BERT}: 1) Link Prediction (LP) component with BERT\textsubscript{BASE} as pre-trained model, 2) LP with fine tuned BERT using with CVE/CWE descriptions (BERT\textsubscript{CVE}), 3) LP with Reconstruction Decoder (RD) using BERT\textsubscript{CVE} as pre-trained model.
\begin{table}[htpb]
\centering
\caption{Prediciton accuracy of LP and RD components with different pre-trained models.}
\resizebox{\linewidth}{!}{
\begin{tabular}{l|ccc|ccc}
\hline
\multicolumn{1}{c|}{\multirow{2}{*}{\textbf{Model}}} & \multicolumn{3}{c|}{\textbf{Test 1 $(k_1, k_2, k_3)$}} & \multicolumn{3}{c}{\textbf{Test 2 $(k_1, k_2, k_3)$}} \\ \cline{2-7} 
\multicolumn{1}{c|}{}                                & \textbf{(1,1,1)} & \textbf{(3,2,1)} & \textbf{(5,2,2)} & \textbf{(1,1,1)} & \textbf{(3,2,1)} & \textbf{(5,2,2)} \\ \hline
LP, BERT\textsubscript{BASE}                              & 0.7829           & 0.8794           & 0.9209           & 0.6995           & 0.8337           & 0.8879                                \\
LP, BERT\textsubscript{CVE}                           & 0.8169           & 0.9137           & 0.9429           & 0.7132           & 0.8505           & 0.9049           
                \\
LP+RD, BERT\textsubscript{CVE} & \textbf{0.8310}  & \textbf{0.9144}  & \textbf{0.9425}  & \textbf{0.7274}  & \textbf{0.8592}  & \textbf{0.9051}  \\ \hline
\end{tabular}
}
\label{tab:pretrained}
\end{table}


\subsection{Training on 1999-2018}
\label{subsec:additionaltraining}
\begin{figure}[htbp]
 \centering
 \includegraphics[width=0.7\linewidth]{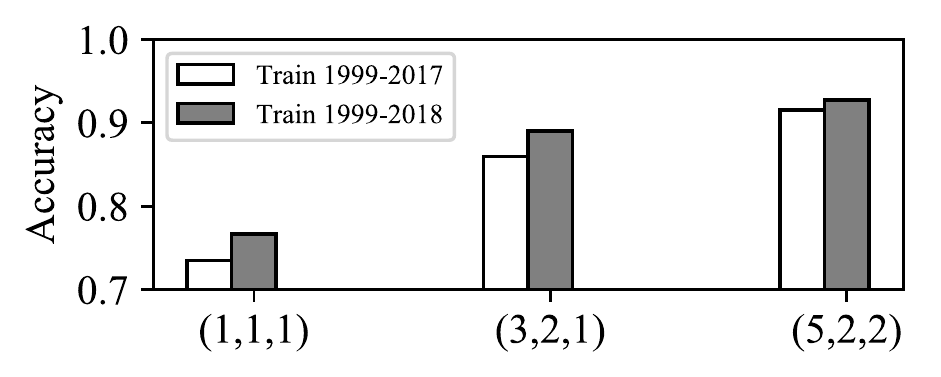}
 \caption{Accuracy of Test 2 before and after adding data from the year 2018 in training.}
 \label{fig:increasing}
\end{figure}

We have performed additional training with CVEs from the year 2018 to predict Test 2 (2019-2020). As expected, recent data improves the performance of the immediate future predictions. 
Fig~\ref{fig:increasing} shows prediction accuracy improvement of \textsc{V2W-BERT} in Test 2 (2019-2020) with additional training data from 2018 and Table~\ref{tab:train2018} shows comparative details.

\begin{table}[htpb]
\caption{Accuracy of Test 2 including 2018 in the training.}
\centering
\resizebox{0.6\linewidth}{!}{
\begin{tabular}{l|rrr}
\hline
\multirow{2}{*}{\textbf{Model}} & \multicolumn{3}{c}{\textbf{Test 2 $(k_1, k_2, k_3)$}}    \\ \cline{2-4} 
                                & \multicolumn{1}{c}{\textbf{(1,1,1)}} & \multicolumn{1}{c}{\textbf{(3,2,1)}} & \multicolumn{1}{c}{\textbf{(5,2,2)}} \\ \hline
Class, TF-IDF NN    & 0.7109 & 0.8444 & 0.8962 \\
Link, TF-IDF NN     & 0.7302 & 0.8636 & 0.9162 \\
Class, BERT\textsubscript{CVE} & 0.7527 & 0.8683 & 0.9090  \\
Link, \textsc{V2W-BERT}         & \textbf{0.7666}                      & \textbf{0.8901}                      & \textbf{0.9273}                        \\ \hline
\end{tabular}
}
\label{tab:train2018}
\end{table}

\subsection{$F_1$-Scores of predicted links} 
\label{subsec:linkpredperformance}
Table~\ref{tab:linkperformance} shows the link prediction performance of the \textsc{V2W-BERT} algorithm and the TF-IDF based link prediction method.
\begin{table}[htpb]
\caption{$F_1$-score of correctly predicted links.}
\centering
\resizebox{0.8\linewidth}{!}{
\begin{tabular}{l|c|c}
\hline
\multirow{2}{*}{\textbf{Model}} & \multicolumn{2}{c}{\textbf{$F_1$-score}}                                                         \\ \cline{2-3} 
                                & \multicolumn{1}{l|}{\textbf{Test 1 (2018)}} & \multicolumn{1}{l}{\textbf{Test 2 (2019-2020)}} \\ \hline
Link, TF-IDF NN & 0.9095          & 0.8816          \\
Link, \textsc{V2W-BERT}       & \textbf{0.9343} & \textbf{0.9156} \\ \hline
\end{tabular}
}
\label{tab:linkperformance}
\vspace{-0.3cm}
\end{table}

\subsection{Predicting a new CWE definition}
\label{subsec:cweprediction}
\begin{figure}[htbp]
 \centering
 \includegraphics[width=\linewidth]{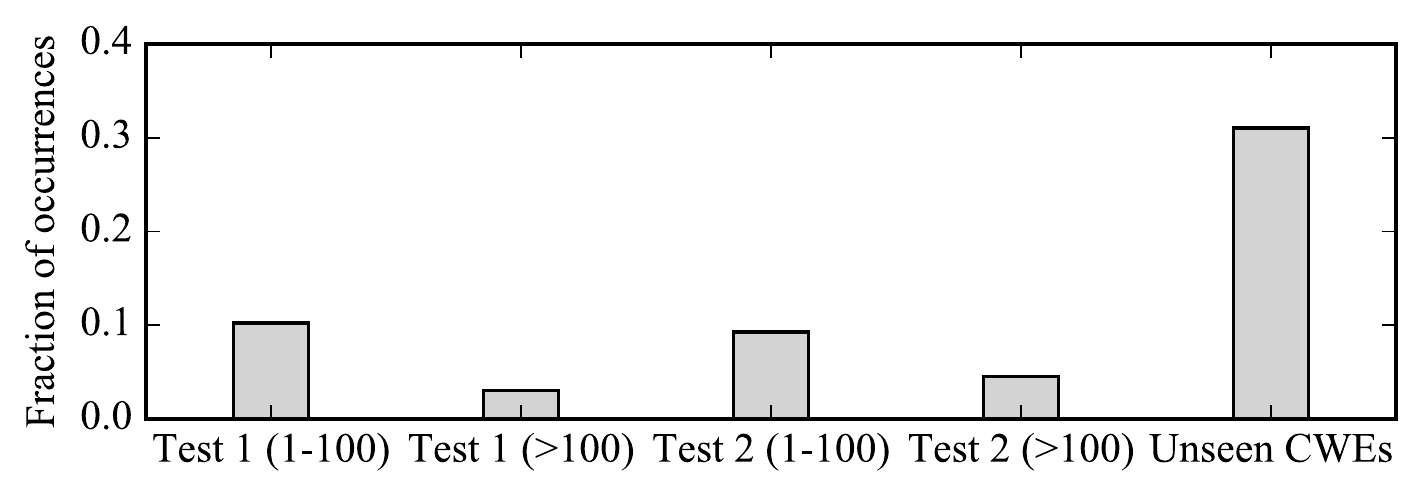}
 \caption{The fraction of occurrences of all unlinks with link threshold set to $\beta=0.90$ in different scenarios.}
 \label{fig:cweprediction}
\end{figure}

Fig~\ref{fig:cweprediction} shows fraction of instances we get all link values less than $\beta=0.90$.
Here ``Test 1 (1-100)'' refers to CVEs associated with CWEs in Test Set 1 with total training instances between 1-100. 
As expected, CVEs of unseen CWEs have the highest fraction of occurrences, because these CVEs have different styles not seen by training method.
Also, the rare type CVEs have higher unlinks to links ratio than frequent ones.
Therefore, if we see only high unlink values to CWEs for some CVE description, we could suggest that experts  take a closer look at the description, and if needed provide a new CWE. 

Table~\ref{tab:predictcwe90} shows how many times we get all link values less than $\beta=0.90$,  and the fraction of such instances. 
We partition the Test sets based on the number of CVEs per CWE class in training. 
\begin{table}[htbp]
\caption{Count of how many times all link values of a CVE to available CWEs are less than $\beta=0.90$ in different scenarios.} 
\centering
\resizebox{0.8\linewidth}{!}{
\begin{tabular}{l|ccc}
\hline
\textbf{Dataset}                 & \textbf{Count} & \textbf{\#Instances} & \textbf{Fraction of} \\ 
                                 &                &                      &
                                 \textbf{Occurrences} \\
\hline
Test 1, 1-100          & 189                   & 1,851                & 0.1021    \\
Test 1, \textgreater 100 & 372                   & 12,236               & 0.0304    \\
Test 2, 1-100            & 357                   & 3,851                & 0.0927    \\
Test 2, \textgreater 100 & 823                   & 18,105               & 0.0454    \\
Unseen CWEs                   & 117                   & 377                 & 0.3103    \\ \hline
\end{tabular}
}
\label{tab:predictcwe90}
\vspace{-0.3cm}
\end{table}

\subsection{Data Augmentation to handle Class Imbalance}
We experimented with data augmentation~\citep{wei2019eda} techniques to handle class imbalance during training. New CVE descriptions are created from the available training CVE descriptions. For CWEs with less than 500 training instances, we gather all text descriptions of the associated CVEs to create a pool of CVE sentences. We take random sentences from the pool of sentences, replace some words with synonyms, and create augmented CVEs description. Table~\ref{tab:augmentation} shows performance comparison before and after the augmentation. Augmentation makes overall convergence faster but achieves similar performance.

\begin{table}[htbp]
\caption{Performance of V2W\textsubscript{BERT} before and after data augmentation.}
\centering
\resizebox{\linewidth}{!}{
\begin{tabular}{l|ccc|ccc}
\hline
\multicolumn{1}{c|}{\multirow{2}{*}{\textbf{Model}}} & \multicolumn{3}{c|}{\textbf{Test 1 $(k_1, k_2, k_3)$}} & \multicolumn{3}{c}{\textbf{Test 2 $(k_1, k_2, k_3)$}} \\ \cline{2-7} 
\multicolumn{1}{c|}{}                                & \textbf{(1,1,1)} & \textbf{(3,2,1)} & \textbf{(5,2,2)} & \textbf{(1,1,1)} & \textbf{(3,2,1)} & \textbf{(5,2,2)} \\ \hline

V2W\textsubscript{BERT}                              & \textbf{0.8362}  &  \textbf{0.914} & \textbf{0.9442}  & 0.7345  & \textbf{0.8594} & \textbf{0.9151} \\ \hline
V2W\textsubscript{BERT}, Aug500                      & 0.8299 & 0.9138	& 0.9425	& \textbf{0.7374}	& 0.8584 &	0.9107 \\ \hline
\end{tabular}
}
\label{tab:augmentation}
\end{table}


\end{document}